\pdfoutput=1

\documentclass[11pt]{article}

\usepackage[final]{coling}

\usepackage{times}
\usepackage{latexsym}

\usepackage[T1]{fontenc}

\usepackage[utf8]{inputenc}

\usepackage{microtype}

\usepackage{inconsolata}

\usepackage{graphicx}
\usepackage{lipsum}
\usepackage{multicol}
\usepackage{multirow}
\usepackage{subcaption}
\usepackage{hyperref}
\title{Multi-ToM: Evaluating Multilingual Theory of Mind Capabilities in Large Language Models}

\author{
    \textbf{Jayanta Sadhu\thanks{Both authors contributed equally}},
    \textbf{Ayan Antik Khan\footnotemark[1]},
    \textbf{Noshin Nawal},
    \textbf{Sanju Basak},\\
    \textbf{Abhik Bhattacharjee},
    \textbf{Rifat Shahriyar}
    \\
    Bangladesh University of Engineering and Technology (BUET)
    \\
    \texttt{\{1705047, 1705036, 1805061, 1805064\}@ugrad.cse.buet.ac.bd,}
    \\
    \texttt{abhik@ra.cse.buet.ac.bd, rifat@cse.buet.ac.bd}
}

\begin{document}
\maketitle

\begin{abstract}
Theory of Mind (ToM) refers to the cognitive ability to infer and attribute mental states to oneself and others. As large language models (LLMs) are increasingly evaluated for social and cognitive capabilities, it remains unclear to what extent these models demonstrate ToM across diverse languages and cultural contexts. In this paper, we introduce a comprehensive study of multilingual ToM capabilities aimed at addressing this gap. Our approach includes two key components: (1) We translate existing ToM datasets into multiple languages, effectively creating a multilingual ToM dataset and (2) We enrich these translations with culturally specific elements to reflect the social and cognitive scenarios relevant to diverse populations. We conduct extensive evaluations of six state-of-the-art LLMs to measure their ToM performance across both the translated and culturally adapted datasets. The results highlight the influence of linguistic and cultural diversity on the models' ability to exhibit ToM, and questions their social reasoning capabilities. This work lays the groundwork for future research into enhancing LLMs' cross-cultural social cognition and contributes to the development of more culturally aware and socially intelligent AI systems. All our data and code are publicly available.\footnote{\href{https://github.com/csebuetnlp/Multi-ToM/}{https://github.com/csebuetnlp/Multi-ToM/}}

\end{abstract}
\section{Introduction}

Theory of Mind (ToM) is a fundamental cognitive skill that allows individuals to understand and attribute mental states—such as beliefs, desires, and intentions—to oneself and others. It plays a critical role in social interaction, enabling humans to predict and interpret behavior based on inferred mental states \cite{zhou2023far}. Traditionally, ToM has been studied within psychology and cognitive science, where it is tested through tasks that assess an individual's ability to understand others' perspectives (e.g., false belief tasks in \citet{wimmer1983beliefs}). However, with the advent of large language models (LLMs), there has been increasing interest in examining whether these AI systems exhibit ToM-like reasoning, i.e., whether they can infer and reason about mental states based on textual inputs \cite{10679907}.

To address these questions, we introduce a multilingual ToM dataset featuring two main components: (a) Direct translations of existing ToM tasks into multiple languages and (b) The integration of culturally specific scenarios. The inclusion of cultural elements aims to assess whether LLMs perform better when provided with data that reflects the social norms and contexts familiar to users from different regions. Since culture plays a vital role in how people interpret mental states and behaviors, evaluating models with culturally nuanced data offers a more comprehensive understanding of their ToM capabilities.

Our key contributions in this work are as follows. (1) We present a multilingual ToM dataset, translated and curated from an existing bilingual ToM dataset, encompassing seven diverse and prominent languages from around the world. We develop an additional dataset that incorporates cultural nuances specific to these languages. (2) We evaluate six state-of-the-art LLMs using both the translated and culturally adapted datasets, investigating whether cultural relevance affects their social reasoning abilities. (3) By examining ToM performance across languages and cultural settings, we provide insights into how the performance of LLMs vary.

\section{Related Works}

Theory of Mind (ToM) is the ability to attribute mental states, such as beliefs and intentions, to oneself and others, aiding in predicting behavior \cite{Premack_Woodruff_1978}. It is essential for social cognition \cite{BARONCOHEN198537}, influencing empathy, relationships \cite{slaughter2002theory}, decision-making \cite{carlson2001individual}, and education \cite{caputi2012longitudinal}.

Recent work \cite{kosinski2023theory, bubeck2023sparks, van2023theory} shows GPT-3.5 and GPT-4 perform well on classical ToM tasks like the false belief task \cite{wimmer1983beliefs}. However, studies reveal their limitations in more complex tasks \cite{ullman2023large, shapira2023clever, jones2023epitome, ma2023tomchallenges, he2023hi, kim2023fantom}. ToMBENCH, an English dataset for benchmarking ToM tasks, was introduced by \citet{chen-etal-2024-tombench}.

Multilingual and cultural aspects are also crucial as \citet{javor2016bilingualism} elaborated that bilinguals have better theory of mind capabilities and empathic skills. \citet{levinson2003space} showed that different languages and cultures shape the way humans perceive and think leveraging the importance of evaluating multilingual and multicultural ToM.

\begin{figure}[h]
    \centering
    \includegraphics[width=\linewidth]{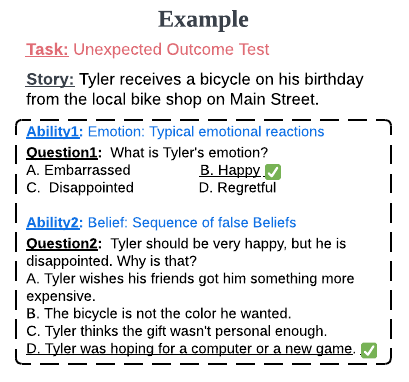}
    \caption{A ToM sample data point consisting of - \\ (a) the type of \textbf{task}, (b) \textbf{a story} capturing a specific scenario, (c) the type of \textbf{ability} being assessed, (d) \textbf{a question} assessing the model's ability to infer the emotions or underlying intentions of a character, (e) multiple \textbf{answer} options providing plausible explanations about the character's actions, with only one being the correct interpretation based on ToM principles.}
    \label{fig:ToM-example}
\end{figure}
\section{Data Preparation}

In this section, we discuss our dataset preparation procedure. We broadly categorize it into three sub-categories. 

\subsection{Data Filtration}
We create the generic MultiToM Dataset by systematically sub-sampling the ToMBench Dataset \citep{chen-etal-2024-tombench}. The original dataset consists of 2,860 samples categorized by 8 tasks and 31 social cognitive abilities in 6 dimensions (ATOMS \citep{beaudoin2020theoryofmind} framework in psychology).

In Figure \ref{fig:ToM-example}, we show an example datapoint. After sampling, our dataset consists of 502 samples covering all categories. We provide the dataset statistics in appendix \ref{appendix:dataset-stats}. Our sampling approach involves two stages.

\subsubsection{Automated Sub-sampling}
For each category, we randomly select a representative baseline example. We then use a sentence transformers model (Appendix \ref{appendix:dataset-generation}) to calculate the semantic distance between the baseline and the remaining samples. Based on the distance, we select examples from various points along the spectrum.

\subsubsection{Post-hoc Human Analysis}
Each category includes examples from diverse social settings (such as school, workplace, family, neighborhood, etc). In order to ensure maximum variation, we carefully make sure that every social scenario is included for an effective ToM evaluation.

\subsection{Data Translation}
We translate all the samples into six languages in addition to English: \textit{Arabic, French, Hindi, Bangla, Russian, Chinese}.
The translation process follows a two-step approach to ensure accuracy - 
\subsubsection{Automatic Translation with Validation}
We formulate an agent-based pipeline with LLMs to translate and auto-validate the quality of translation. Initially, each sample is translated using GPT-4 (\textit{Agent-1}). The quality of the translation is then validated by GPT-3.5 (\textit{Agent-2}), which provides a feedback. The feedback along with the original sample is provided to GPT-3.5 (\textit{Agent-3}) again. Any necessary revisions are made based on this feedback before finalizing the translated sample. The pipeline is demonstrated in Figure \ref{fig:ToM-translation}. 
\subsubsection{Human-in-the-loop Verification}
Once the dataset has been fully translated, we employ a human-in-the-loop verification process. This involves manually back-translating a significant portion of the samples using Google Translate to assess the translation quality. We ensure that the translated samples preserve the core narrative of the original data.
\subsection{Cultural Element Induction}
After generating the generic dataset samples, we introduce specific cultural nuances reflective of the cultures in which our selected subset of languages are spoken.
\subsubsection{Motivation}
Theory of Mind (ToM) is a social cognitive ability, involving the capacity to infer others' mental states. As social interactions are shaped by cultural contexts, it is essential to incorporate cultural nuances into ToM evaluations. Factors like language development and social environments can significantly influence ToM performance. For instance, \citet{Shahaeian2011} argue that children in collectivist cultures tend to develop certain aspects of ToM earlier than those in individualistic cultures, which highlights the role of cultural differences in ToM abilities. 
\begin{figure}[h]
    \centering
    \includegraphics[width=\linewidth]{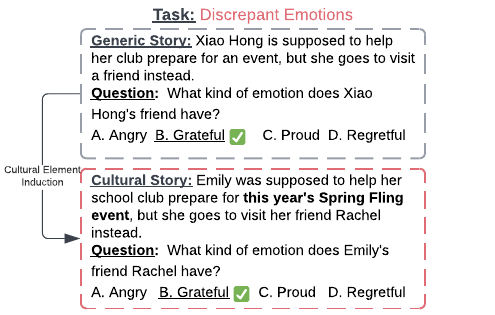}
    \caption{Process of cultural element induction in a discrepant emotions task. The generic story is culturally adapted to reflect a Western context. Despite the cultural modifications, the core narrative remains unchanged.}
    \label{fig:generic-cultural}
\end{figure}
\subsubsection{Procedure}
We utilize Llama-3.1 and GPT-3.5 Turbo to incorporate specific cultural settings or modify cultural elements within the stories of the data samples. While adapting these aspects, we ensure intact core narrative. Similarly, we adapt the \textit{Question} and the \textit{Options} to align with the revised cultural context, ensuring that the fundamental analysis remains unaffected. Additionally, we perform post-hoc human validation to confirm that the core narrative remains preserved in the culturally adapted data samples. (Appendix \ref{appendix:cultural-nuance-injection})


\section{Experiments}

\subsection{Experimental Setup}
We evaluated a total of six popular LLMs, which include Claude-3.5-Sonnet \cite{claude35sonnet}, Claude Instant v1.2 \cite{anthropic2023claude}, GPT-4o \cite{achiam2023gpt}, GPT-3.5-Turbo \cite{openai2023gpt35}, Llama-3.1-8b Instruct \cite{llama3modelcard} and Llama-2-7b Chat \cite{touvron2023llama}. For all open and closed LLMs, we strictly abide by their terms and get access through their official model weights and APIs respectively.

\subsection{Prompting}
For each data point, we prompt the model with direct questions that includes the story, related question and options to choose from. The options of the question is randomly shuffled to avoid any selection bias \cite{xue-etal-2024-strengthened, zheng2024largelanguagemodelsrobust}. The system instruction templates do not change for different languages, either for directly translated questions or culturally nuanced variations.

\section{Results and Evaluation}

\subsection{Performance Analysis}
\label{subsection:performance_analysis}
We provide a qualitative analysis of the LLMs in different ToM tasks and emotions in this section. The detailed results on which this analysis is based on is given in appendix \ref{appendix:result-details}.

\paragraph{Differences Across ToM Tasks:}
We observe that the Faux-pas Recognition Test (FRT) shows relatively high performance due to its simpler true/false format, reducing task complexity. In contrast, the Scalar Implicature Task (SIT) performs poorly, likely due to its reliance on mathematical reasoning, an area where LLMs are limited. Performance across tasks remains consistent across different languages, suggesting that LLMs' ability to perform ToM tasks is more influenced by the task's nature than by the language used.

\paragraph{Differences Across ToM Abilities:}
We observe similar task effects across abilities, with abilities maintaining their relative positioning across languages. Non-literal Communication (NLC) is generally the strongest, likely due to its overlap with FRT. In contrast, knowledge-related tasks perform poorly, as they require understanding that others have different knowledge based on perceptions or information, a known weakness of LLMs.

\begin{figure*}[t]
    \centering
    \begin{subfigure}[b]{0.48\textwidth}
        \centering
        \includegraphics[width=\textwidth]{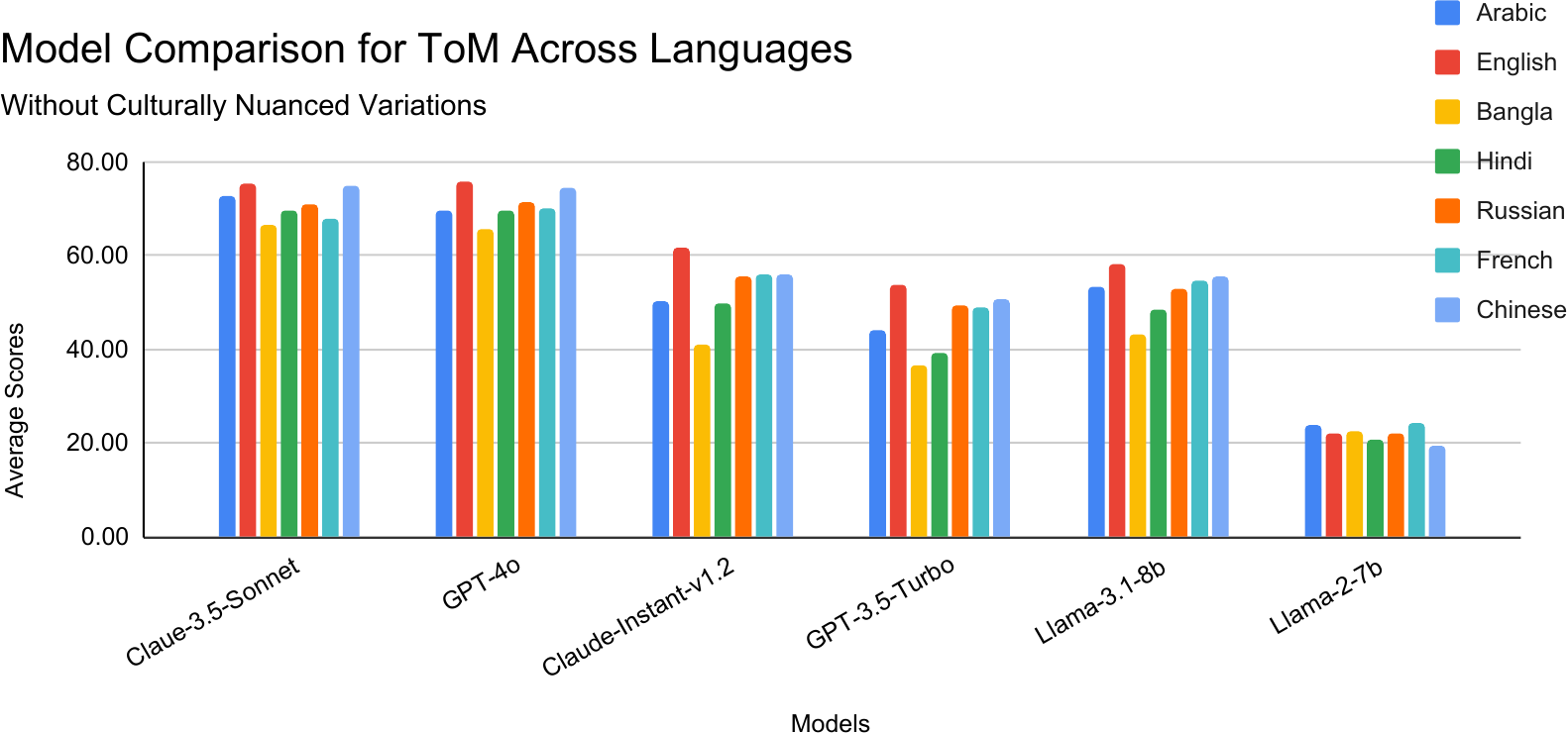}
        \caption{Average score comparison of models for each language without culturally nuanced dataset.}
        \label{subfig:model_comparison_without_culture}
    \end{subfigure}
    \hfill
    \begin{subfigure}[b]{0.48\textwidth}
        \centering
        \includegraphics[width=\textwidth]{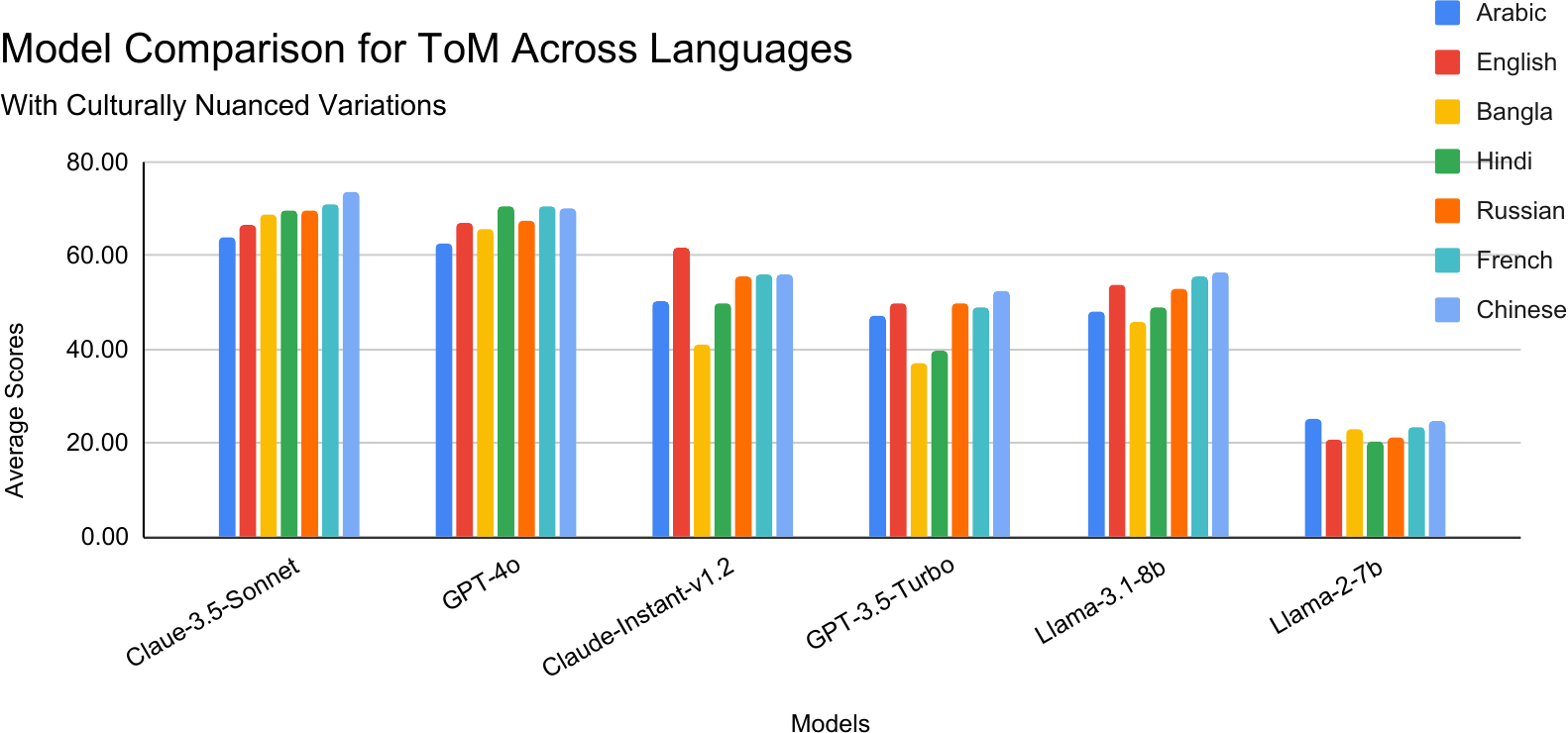}
        \caption{Average score comparison of models for each language with culturally nuanced dataset.}
        \label{subfig:model_comparison_with_culture}
    \end{subfigure}
    \begin{subfigure}[b]{0.48\textwidth}
        \centering
        \includegraphics[width=\textwidth]{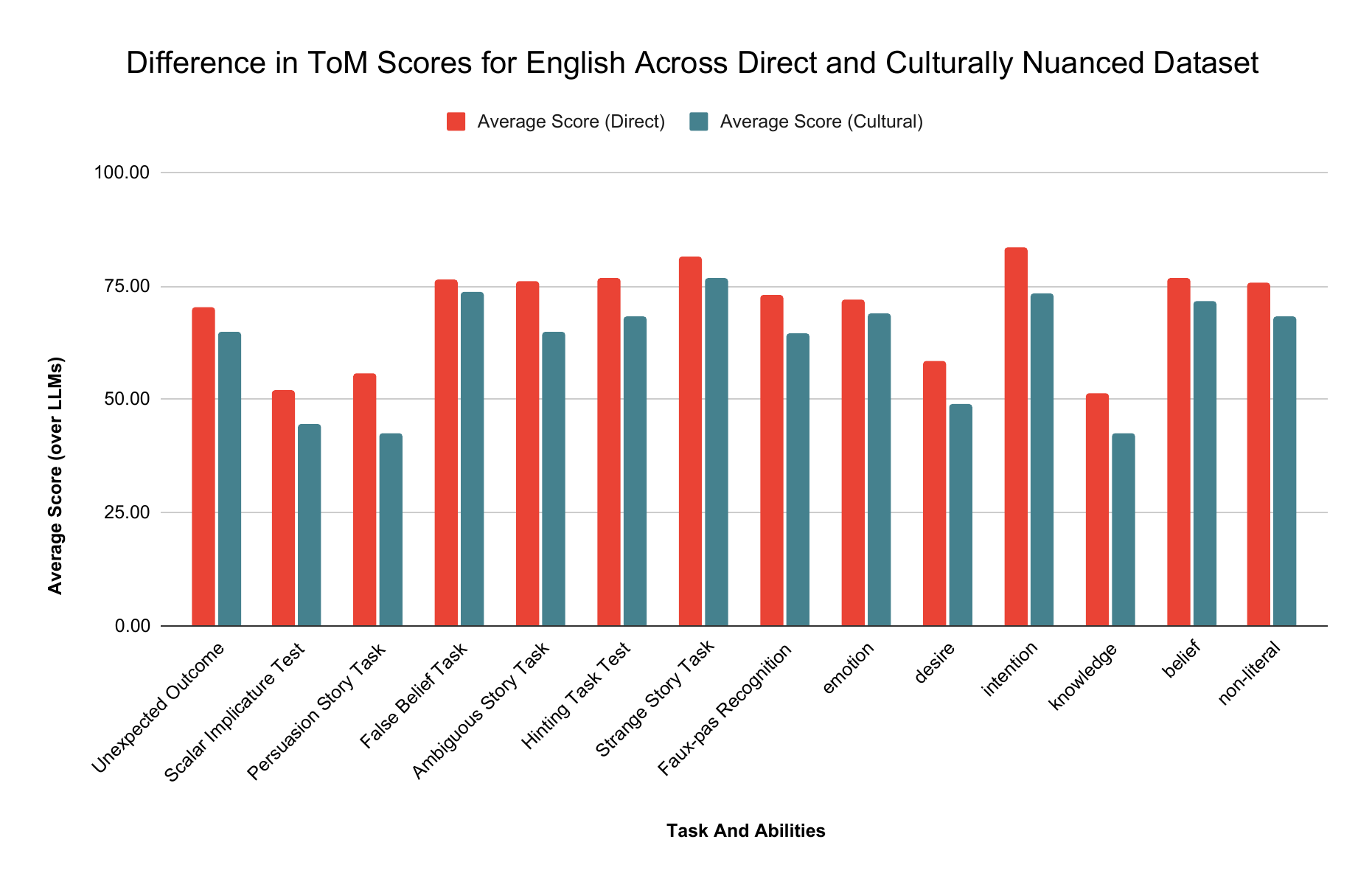}
        \caption{ToM Score comparisons for tasks and abilities for plain and culturally nuanced dataset for English.}
        \label{subfig:english_task_comparison}
    \end{subfigure}
    \begin{subfigure}[b]{0.48\textwidth}
        \centering
        \includegraphics[width=\textwidth]{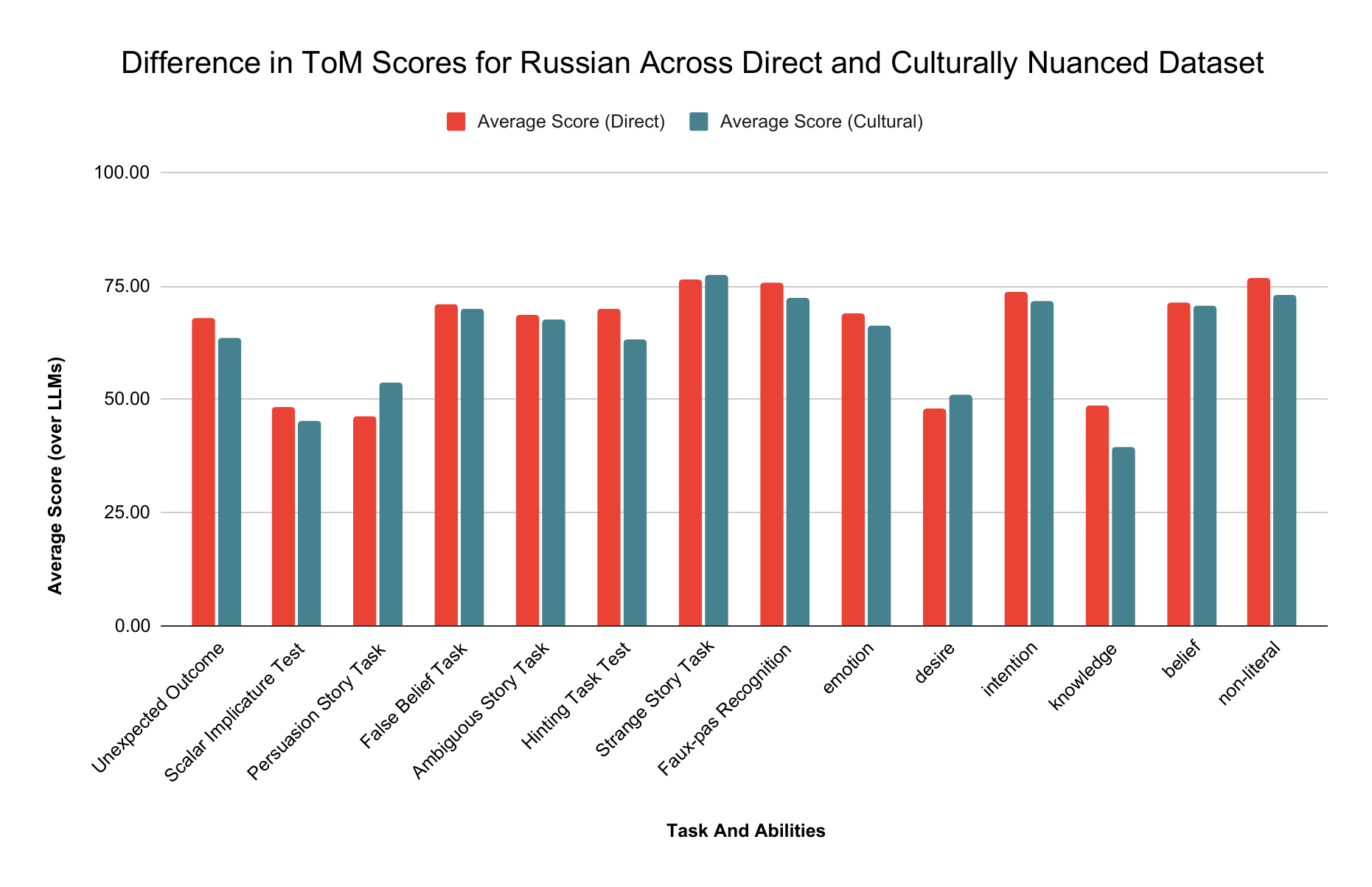}
        \caption{ToM Score comparisons for tasks and abilities for plain and culturally nuanced dataset for Russian.}
        \label{subfig:russian_task_comparison}
    \end{subfigure}
    \caption{Comparative analysis between LLM's abilities on ToM tasks and abilities for different setup. \textit{(Note that the average scores for each task and abilities in Figure \ref{subfig:english_task_comparison} and \ref{subfig:russian_task_comparison} are averaged over the top three performing LLMs: Claude-3.5-Sonnet, GPT-4o and LLama-3.1-8b)}}
\end{figure*}


\subsection{Fine-grained Analysis}
Comparing Figures \ref{subfig:model_comparison_without_culture} and \ref{subfig:model_comparison_with_culture}, we observe that average LLM performance across languages does not differ significantly between direct translations and culturally adapted datasets, though it varies by language. LLMs perform better in resource-rich languages like \textbf{English}, \textbf{Chinese}, and \textbf{French}, while low-resource languages such as \textbf{Bangla}, \textbf{Hindi}, and \textbf{Arabic} score lower. As discussed in subsection \ref{subsection:performance_analysis}, challenging ToM tasks pose similar difficulties across all languages and models. Thus, we conclude that LLM cognitive performance is influenced by the linguistic resources available. 

In Figures \ref{subfig:english_task_comparison} and \ref{subfig:russian_task_comparison}, we present the variation in task and ability scores for English and Russian. In both cases, directly translated prompts generally outperform culturally nuanced datasets, with minimal changes observed in Russian, with few exceptions. However, this trend is more pronounced in English, invoking further analysis.

For the English language, we conduct a detailed examination of cases where an LLM answers correctly on the directly translated dataset but incorrectly on the culturally nuanced data. Our investigation reveals that the inclusion of culturally specific elements, such as names of people, places, and irrelevant details in the questions and options, likely disrupt the model's reasoning process. These additional elements, while culturally relevant, may have introduced noise or complexity that distract the LLM from focusing on the core elements of the task.
This suggests that the cognitive abilities of LLMs can be sensitive to subtle linguistic variations, including extraneous or culturally specific information. We provide some examples in appendix \ref{appendix:incoherence-tests} for clarification.
\section{Conclusion}
In this study, we introduce the first multilingual dataset for evaluating Theory of Mind (ToM) abilities in LLMs and establish benchmarks for six popular models. To gain deeper insights into LLM performance, we conduct experiments using both directly translated and culturally adapted datasets. Our findings on the impact of cultural variations suggest that future research should explore the reasoning capabilities of LLMs, focusing on rephrasing, paraphrasing, and the addition of irrelevant content
and targeted element modification in ToM tasks.
\clearpage
\section*{Limitations}
We discuss the limitations of our work as follows. \\~\\
\textbf{Dataset Size} For this study, we sub-sample the original ToM dataset to approximately one-sixth of its size. Since we choose seven different languages and two different test cases(Generic and Cultural), this approach was necessary due to resource constraints. Although we made efforts to preserve the original ratio in each of the 8 tasks and 6 emotions, the sub-sampling process resulted in a reduced dataset. In future, a more comprehensive study could be conducted using the full dataset, eliminating the need for sub-sampling. Furthermore, during the construction of the culturally nuanced dataset, future work could experiment with several stages and methods of nuance induction. Through which, a more robust and comprehensive dataset could be developed. However, due to the challenges involved and resource limitations, we were unable to fully explore these improvements in the current study.
\\~\\
\textbf{Models \& Prompting Methods} We use six state-of-the-art LLMs for ToM evaluation. Future work could include more open-source LLMs since we believe open source LLMs have the capability to serve a wider user-base. We use Vanilla Prompting method for inference task, we ignore CoT (Chain-of-Thought) prompting based on the findings on \citet{chen-etal-2024-tombench}, they perform the study on both Vanilla and CoT prompting and results indicate that CoT prompting almost never improves ToM performance and at times leads to decline. Still it is an area of exploration for the \textit{culturally nuanced dataset}. We leave the newer prompting methods such as \textit{perspective taking} \citep{wilf-etal-2024-think} and \textit{foresee and reflect} \citep{zhou2023farlargelanguagemodels} for future exploration.
\\~\\
\textbf{Language Barriers}
When injecting cultural nuances, we maintain the use of English to simplify the verification process and ensure that the core narrative of each data sample remains consistent. Although the authors are familiar with more than one of the languages used in this study, it would be interesting to explore injecting cultural nuances directly in those specific languages. This step would require the involvement of individuals proficient in multiple languages for accurate verification. We identify this as an area for future exploration.

\section{Ethical Considerations}
We discuss potential ethical considerations as follows. \\~\\
\textbf{Cultural Sensitivity and Representation}
While injecting cultural nuances, we believe that it is crucial to ensure that the stories and scenarios reflect diverse cultures accurately and respectfully. We manually evaluate each of the culturally nuanced data samples to prevent negative stereotypes or harmful cultural biases. Still, it is important to note that cultural stereotypes could be presented in the form of cultural nuance injection. 
\\~\\
\textbf{LLMs and Humans}
As we explore and enhance LLMs' ability to simulate human-like reasoning, our goal is to improve their capacity to interpret and understand human mental states. It is important to emphasize that we do not aim to humanize LLMs. Rather, we seek to bridge the communication gap between machines and humans.


\bibliography{custom}

\begin{thebibliography}{34}
\providecommand{\natexlab}[1]{#1}

\bibitem[{Achiam et~al.(2023)Achiam, Adler, Agarwal, Ahmad, Akkaya, Aleman, Almeida, Altenschmidt, Altman, Anadkat et~al.}]{achiam2023gpt}
Josh Achiam, Steven Adler, Sandhini Agarwal, Lama Ahmad, Ilge Akkaya, Florencia~Leoni Aleman, Diogo Almeida, Janko Altenschmidt, Sam Altman, Shyamal Anadkat, et~al. 2023.
\newblock Gpt-4 technical report.
\newblock \emph{arXiv preprint arXiv:2303.08774}.

\bibitem[{AI@Meta(2024)}]{llama3modelcard}
AI@Meta. 2024.
\newblock \href {https://github.com/meta-llama/llama3/blob/main/MODEL_CARD.md} {Llama 3 model card}.

\bibitem[{Anthropic(2023{\natexlab{a}})}]{claude35sonnet}
Anthropic. 2023{\natexlab{a}}.
\newblock \href {https://www.anthropic.com/news/claude-3-5-sonnet} {Claude 3.5 sonnet announcement}.
\newblock Accessed: 2024-09-17.

\bibitem[{Anthropic(2023{\natexlab{b}})}]{anthropic2023claude}
Anthropic. 2023{\natexlab{b}}.
\newblock Claude instant v1.2.
\newblock \url{https://www.anthropic.com}.

\bibitem[{Baron-Cohen et~al.(1985)Baron-Cohen, Leslie, and Frith}]{BARONCOHEN198537}
Simon Baron-Cohen, Alan~M. Leslie, and Uta Frith. 1985.
\newblock \href {https://doi.org/10.1016/0010-0277(85)90022-8} {Does the autistic child have a “theory of mind” ?}
\newblock \emph{Cognition}, 21(1):37--46.

\bibitem[{Beaudoin et~al.(2020)Beaudoin, Émilie Leblanc, Gagner, and Beauchamp}]{beaudoin2020theoryofmind}
Catherine Beaudoin, Émilie Leblanc, Claudia Gagner, and Miriam~H. Beauchamp. 2020.
\newblock \href {https://doi.org/10.3389/fpsyg.2019.02905} {Systematic review and inventory of theory of mind measures for young children}.
\newblock \emph{Frontiers in Psychology}, 10:2905.

\bibitem[{Bhattacharjee et~al.(2022)Bhattacharjee, Hasan, Ahmad, Mubasshir, Islam, Iqbal, Rahman, and Shahriyar}]{bhattacharjee-etal-2022-banglabert}
Abhik Bhattacharjee, Tahmid Hasan, Wasi Ahmad, Kazi~Samin Mubasshir, Md~Saiful Islam, Anindya Iqbal, M.~Sohel Rahman, and Rifat Shahriyar. 2022.
\newblock \href {https://aclanthology.org/2022.findings-naacl.98} {{B}angla{BERT}: Language model pretraining and benchmarks for low-resource language understanding evaluation in {B}angla}.
\newblock In \emph{Findings of the Association for Computational Linguistics: NAACL 2022}, pages 1318--1327, Seattle, United States. Association for Computational Linguistics.

\bibitem[{Bubeck et~al.(2023)Bubeck, Chandrasekaran, Eldan, Gehrke, Horvitz, Kamar, Lee, Lee, Li, Lundberg et~al.}]{bubeck2023sparks}
S{\'e}bastien Bubeck, Varun Chandrasekaran, Ronen Eldan, Johannes Gehrke, Eric Horvitz, Ece Kamar, Peter Lee, Yin~Tat Lee, Yuanzhi Li, Scott Lundberg, et~al. 2023.
\newblock Sparks of artificial general intelligence: Early experiments with gpt-4.
\newblock \emph{arXiv preprint arXiv:2303.12712}.

\bibitem[{Caputi et~al.(2012)Caputi, Lecce, Pagnin, and Banerjee}]{caputi2012longitudinal}
Marcella Caputi, Serena Lecce, Adriano Pagnin, and Robin Banerjee. 2012.
\newblock Longitudinal effects of theory of mind on later peer relations: the role of prosocial behavior.
\newblock \emph{Developmental psychology}, 48(1):257.

\bibitem[{Carlson and Moses(2001)}]{carlson2001individual}
Stephanie~M Carlson and Louis~J Moses. 2001.
\newblock Individual differences in inhibitory control and children's theory of mind.
\newblock \emph{Child development}, 72(4):1032--1053.

\bibitem[{Chen et~al.(2024)Chen, Wu, Zhou, Wen, Bi, Jiang, Cao, Hu, Lai, Xiong, and Huang}]{chen-etal-2024-tombench}
Zhuang Chen, Jincenzi Wu, Jinfeng Zhou, Bosi Wen, Guanqun Bi, Gongyao Jiang, Yaru Cao, Mengting Hu, Yunghwei Lai, Zexuan Xiong, and Minlie Huang. 2024.
\newblock \href {https://aclanthology.org/2024.acl-long.847} {{T}o{MB}ench: Benchmarking theory of mind in large language models}.
\newblock In \emph{Proceedings of the 62nd Annual Meeting of the Association for Computational Linguistics (Volume 1: Long Papers)}, pages 15959--15983, Bangkok, Thailand. Association for Computational Linguistics.

\bibitem[{Hasan et~al.(2020)Hasan, Bhattacharjee, Samin, Hasan, Basak, Rahman, and Shahriyar}]{hasan-etal-2020-low}
Tahmid Hasan, Abhik Bhattacharjee, Kazi Samin, Masum Hasan, Madhusudan Basak, M.~Sohel Rahman, and Rifat Shahriyar. 2020.
\newblock \href {https://doi.org/10.18653/v1/2020.emnlp-main.207} {Not low-resource anymore: Aligner ensembling, batch filtering, and new datasets for {B}engali-{E}nglish machine translation}.
\newblock In \emph{Proceedings of the 2020 Conference on Empirical Methods in Natural Language Processing (EMNLP)}, pages 2612--2623, Online. Association for Computational Linguistics.

\bibitem[{He et~al.(2023)He, Wu, Jia, Mihalcea, Chen, and Deng}]{he2023hi}
Yinghui He, Yufan Wu, Yilin Jia, Rada Mihalcea, Yulong Chen, and Naihao Deng. 2023.
\newblock Hi-tom: A benchmark for evaluating higher-order theory of mind reasoning in large language models.
\newblock \emph{arXiv preprint arXiv:2310.16755}.

\bibitem[{Javor and Javor(2016)}]{javor2016bilingualism}
Rebeka Javor and R~Javor. 2016.
\newblock Bilingualism, theory of mind and perspective-taking: The effect of early bilingual exposure.
\newblock \emph{Psychology and Behavioral Sciences}, 5(6):143--148.

\bibitem[{Jones et~al.(2023)Jones, Trott, and Bergen}]{jones2023epitome}
Cameron~Robert Jones, Sean Trott, and Ben Bergen. 2023.
\newblock Epitome: Experimental protocol inventory for theory of mind evaluation.
\newblock In \emph{First Workshop on Theory of Mind in Communicating Agents}.

\bibitem[{Kim et~al.(2023)Kim, Sclar, Zhou, Bras, Kim, Choi, and Sap}]{kim2023fantom}
Hyunwoo Kim, Melanie Sclar, Xuhui Zhou, Ronan~Le Bras, Gunhee Kim, Yejin Choi, and Maarten Sap. 2023.
\newblock Fantom: A benchmark for stress-testing machine theory of mind in interactions.
\newblock \emph{arXiv preprint arXiv:2310.15421}.

\bibitem[{Kosinski(2023)}]{kosinski2023theory}
Michal Kosinski. 2023.
\newblock Theory of mind might have spontaneously emerged in large language models.
\newblock \emph{arXiv preprint arXiv:2302.02083}.

\bibitem[{Levinson(2003)}]{levinson2003space}
Stephen~C Levinson. 2003.
\newblock \emph{Space in language and cognition: Explorations in cognitive diversity}, volume~5.
\newblock Cambridge University Press.

\bibitem[{Ma et~al.(2023)Ma, Gao, and Xu}]{ma2023tomchallenges}
Xiaomeng Ma, Lingyu Gao, and Qihui Xu. 2023.
\newblock Tomchallenges: A principle-guided dataset and diverse evaluation tasks for exploring theory of mind.
\newblock \emph{arXiv preprint arXiv:2305.15068}.

\bibitem[{Mao et~al.(2024)Mao, Liu, Ni, Lin, and He}]{10679907}
Yuanyuan Mao, Shuang Liu, Qin Ni, Xin Lin, and Liang He. 2024.
\newblock \href {https://doi.org/10.1109/TCSS.2024.3416707} {A review on machine theory of mind}.
\newblock \emph{IEEE Transactions on Computational Social Systems}, pages 1--19.

\bibitem[{OpenAI(2023)}]{openai2023gpt35}
OpenAI. 2023.
\newblock \href {https://openai.com/research/gpt-3-5} {Gpt-3.5 turbo}.
\newblock Accessed: 2024-09-17.

\bibitem[{Premack and Woodruff(1978)}]{Premack_Woodruff_1978}
David Premack and Guy Woodruff. 1978.
\newblock \href {https://doi.org/10.1017/S0140525X00076512} {Does the chimpanzee have a theory of mind?}
\newblock \emph{Behavioral and Brain Sciences}, 1(4):515–526.

\bibitem[{Shahaeian et~al.(2011)Shahaeian, Peterson, Slaughter, and Wellman}]{Shahaeian2011}
Ameneh Shahaeian, Candida~C. Peterson, Virginia Slaughter, and Henry~M. Wellman. 2011.
\newblock \href {https://doi.org/10.1037/a0023899} {Culture and the sequence of steps in theory of mind development}.
\newblock \emph{Developmental Psychology}, 47(5):1239--1247.

\bibitem[{Shapira et~al.(2023)Shapira, Levy, Alavi, Zhou, Choi, Goldberg, Sap, and Shwartz}]{shapira2023clever}
Natalie Shapira, Mosh Levy, Seyed~Hossein Alavi, Xuhui Zhou, Yejin Choi, Yoav Goldberg, Maarten Sap, and Vered Shwartz. 2023.
\newblock Clever hans or neural theory of mind? stress testing social reasoning in large language models.
\newblock \emph{arXiv preprint arXiv:2305.14763}.

\bibitem[{Slaughter et~al.(2002)Slaughter, Dennis, and Pritchard}]{slaughter2002theory}
Virginia Slaughter, Michelle~J Dennis, and Michelle Pritchard. 2002.
\newblock Theory of mind and peer acceptance in preschool children.
\newblock \emph{British journal of developmental psychology}, 20(4):545--564.

\bibitem[{Touvron et~al.(2023)Touvron, Martin, Stone, Albert, Almahairi, Babaei, Bashlykov, Batra, Bhargava, Bhosale et~al.}]{touvron2023llama}
Hugo Touvron, Louis Martin, Kevin Stone, Peter Albert, Amjad Almahairi, Yasmine Babaei, Nikolay Bashlykov, Soumya Batra, Prajjwal Bhargava, Shruti Bhosale, et~al. 2023.
\newblock Llama 2: Open foundation and fine-tuned chat models.
\newblock \emph{arXiv preprint arXiv:2307.09288}.

\bibitem[{Ullman(2023)}]{ullman2023large}
Tomer Ullman. 2023.
\newblock Large language models fail on trivial alterations to theory-of-mind tasks.
\newblock \emph{arXiv preprint arXiv:2302.08399}.

\bibitem[{van Duijn et~al.(2023)van Duijn, van Dijk, Kouwenhoven, de~Valk, Spruit, and van~der Putten}]{van2023theory}
Max~J van Duijn, Bram van Dijk, Tom Kouwenhoven, Werner de~Valk, Marco~R Spruit, and Peter van~der Putten. 2023.
\newblock Theory of mind in large language models: Examining performance of 11 state-of-the-art models vs. children aged 7-10 on advanced tests.
\newblock \emph{arXiv preprint arXiv:2310.20320}.

\bibitem[{Wilf et~al.(2024)Wilf, Lee, Liang, and Morency}]{wilf-etal-2024-think}
Alex Wilf, Sihyun Lee, Paul~Pu Liang, and Louis-Philippe Morency. 2024.
\newblock \href {https://aclanthology.org/2024.acl-long.451} {Think twice: Perspective-taking improves large language models{'} theory-of-mind capabilities}.
\newblock In \emph{Proceedings of the 62nd Annual Meeting of the Association for Computational Linguistics (Volume 1: Long Papers)}, pages 8292--8308, Bangkok, Thailand. Association for Computational Linguistics.

\bibitem[{Wimmer and Perner(1983)}]{wimmer1983beliefs}
Heinz Wimmer and Josef Perner. 1983.
\newblock Beliefs about beliefs: Representation and constraining function of wrong beliefs in young children's understanding of deception.
\newblock \emph{Cognition}, 13(1):103--128.

\bibitem[{Xue et~al.(2024)Xue, Hu, Liu, Liao, Li, Han, Zhao, and Yin}]{xue-etal-2024-strengthened}
Mengge Xue, Zhenyu Hu, Liqun Liu, Kuo Liao, Shuang Li, Honglin Han, Meng Zhao, and Chengguo Yin. 2024.
\newblock \href {https://aclanthology.org/2024.acl-long.237} {Strengthened symbol binding makes large language models reliable multiple-choice selectors}.
\newblock In \emph{Proceedings of the 62nd Annual Meeting of the Association for Computational Linguistics (Volume 1: Long Papers)}, pages 4331--4344, Bangkok, Thailand. Association for Computational Linguistics.

\bibitem[{Zheng et~al.(2024)Zheng, Zhou, Meng, Zhou, and Huang}]{zheng2024largelanguagemodelsrobust}
Chujie Zheng, Hao Zhou, Fandong Meng, Jie Zhou, and Minlie Huang. 2024.
\newblock \href {https://arxiv.org/abs/2309.03882} {Large language models are not robust multiple choice selectors}.
\newblock \emph{Preprint}, arXiv:2309.03882.

\bibitem[{Zhou et~al.(2023{\natexlab{a}})Zhou, Madaan, Potharaju, Gupta, McKee, Holtzman, Pujara, Ren, Mishra, Nematzadeh, Upadhyay, and Faruqui}]{zhou2023farlargelanguagemodels}
Pei Zhou, Aman Madaan, Srividya~Pranavi Potharaju, Aditya Gupta, Kevin~R. McKee, Ari Holtzman, Jay Pujara, Xiang Ren, Swaroop Mishra, Aida Nematzadeh, Shyam Upadhyay, and Manaal Faruqui. 2023{\natexlab{a}}.
\newblock \href {https://arxiv.org/abs/2310.03051} {How far are large language models from agents with theory-of-mind?}
\newblock \emph{Preprint}, arXiv:2310.03051.

\bibitem[{Zhou et~al.(2023{\natexlab{b}})Zhou, Madaan, Potharaju, Gupta, McKee, Holtzman, Pujara, Ren, Mishra, Nematzadeh et~al.}]{zhou2023far}
Pei Zhou, Aman Madaan, Srividya~Pranavi Potharaju, Aditya Gupta, Kevin~R McKee, Ari Holtzman, Jay Pujara, Xiang Ren, Swaroop Mishra, Aida Nematzadeh, et~al. 2023{\natexlab{b}}.
\newblock How far are large language models from agents with theory-of-mind?
\newblock \emph{arXiv preprint arXiv:2310.03051}.

\end{thebibliography}

\clearpage 
\appendix
\section*{Appendix}

\section{Data Generation Pipeline}
\label{appendix:dataset-generation}
\begin{figure}[h]
    \centering
    \includegraphics[width=\linewidth]{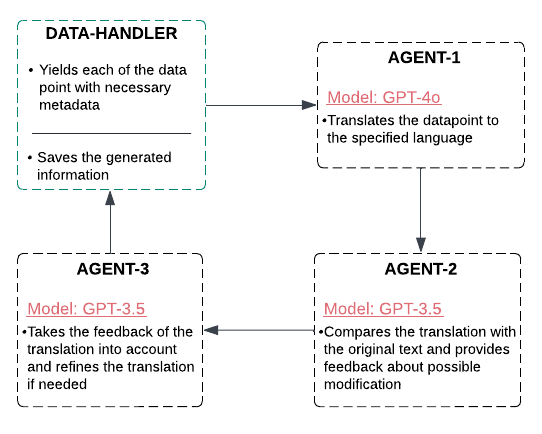}
    \caption{Multi-step translation process for MultiToM. The \textit{Data-Handler} provides each data point along with necessary metadata. \textit{Agent-1} (GPT-4) translates the data point into the specified language. \textit{Agent-2} (GPT-3.5) reviews the translation, comparing it with the original text and suggesting possible modifications. Finally, \textit{Agent-3} (GPT-3.5) refines the translation based on the feedback, and the \textit{Data-Handler} saves the final version.}
    \label{fig:ToM-translation}
\end{figure}

In the Data Filtration step, we first use an automated sub-sampling procedure where we use an open-source sentence transformer model\footnote{\url{https://huggingface.co/sentence-transformers/all-MiniLM-L6-v2}}, we choose the model based on open-source credibility and ease of use. During the data filtration, we aim to minimize the samples in a way such that variation and distribution of the amount of different abilities and tasks remain consistent with the original dataset. The need to filter the original dataset stems from the constraint in computational resources.
\\~\\
We present the automated data translation pipeline in Figure \ref{fig:ToM-translation}. We include the prompts for each Agent in the translation pipeline in Table \ref{tab:translation-prompts}.
\section{Dataset Statistics}
\label{appendix:dataset-stats}
Table \ref{tab:generic-cultural-stats} includes the Statistics of our sampled Dataset based on 8 tasks and 6 dimensions. A noticeable characteristic of the statistics is that the Average Story Length increases in the culturally nuanced versions of the dataset, due to newer cultural elements being introduced. In order to extract the \textit{ASL} (Average Story Length) Characteristic, we use tokenizers. For Chinese language, we use \textit{jieba}\footnote{\url{https://github.com/fxsjy/jieba}}, which is a Chinese text segmentation tool. For Bengali language, we use a normalizer introduced in  \citet{hasan-etal-2020-low}, and then we use the tokenizer for BanglaBERT \citep{bhattacharjee-etal-2022-banglabert}, for all other languages, we use the tokenizer for xlm-roberta-base\footnote{\url{https://huggingface.co/FacebookAI/xlm-roberta-base}}. 
\begin{table*}
    \centering
    \begin{minipage}{\textwidth}
    \begin{tabular}{|p{16cm}|}
    \hline
    \multicolumn{1}{|c|}{\textbf{System Prompt for Agent-1 (GPT-4o)}} \\
    \hline 
    You are an AI assistant whose job is to translate some given questions from 
    \{source\_language\} to \{destination\_language\}. While translation, there will be 
    tags starting with \#\#\# signs (like \#\#\#Question). Do not translate these tags. 
    Make sure you translate all the tag contents. Please follow the guidelines: \\~\\
    
    1. Maintain Meaning: Ensure the translated question conveys the original intent. \\
    2. Cultural Adaptation: Adjust cultural references to suit the target language. For  example, adapt place names, idioms, festivals, toys, objects and cultural symbols as 
    needed (e.g., 'Louvre Museum' should be localized appropriately). \\
    3. Context Sensitivity: Choose translations that match the context, avoiding direct word-for-word translations that may distort meaning. \\
    4. Natural Expression: Ensure the translation flows naturally in \{destination\_language\}, preserving the readability and coherence. \\ 
    5. There will be options with A,B,C or D. Do not translate the options letters and keep them in the same order. \\
    6. Make sure all the elements are present in your response, like \#\#\#STORY, \#\#\#QUESTION and \#\#\#OPTIONS. \\
    \hline
    \end{tabular}
    \end{minipage}

    \vspace{1cm}
    \begin{minipage}{\textwidth}
    \begin{tabular}{|p{16cm}|}
    \hline
    \multicolumn{1}{|c|}{\textbf{System Prompt for Agent-2 (GPT-3.5)}} \\
    \hline 
    You are tasked with verifying a translation from \{source\_language\} to \{destination\_language\}. You will be given the original text and the translated text. Your job is to check the following: \\

    1. Accuracy of Meaning: Ensure that the translated text preserves the same meaning as the original. Point out any inconsistencies or loss of information. \\
    2. Cultural Adaptation: Verify if any cultural references or context-specific terms are correctly translated, maintaining cultural appropriateness for \{destination\_language\}. \\ 
    3. Contextual Relevance: Check if the translation uses contextually correct words or phrases, ensuring that any ambiguities or multiple meanings are handled properly. \\
    4. Suggestions: Provide constructive feedback on how to improve the translation, if necessary, in English. \\
    Your goal is to ensure that the translation is accurate, natural, and culturally appropriate. Your task is to return two tags in your output:
    \#\#\#Quality: respond with okay if the translation looks good.
    \#\#\#Feedback: Only respond with feedback if the translation is not good. Put this tag before the feedback. \\
    Example:
    1. \#\#\#Quality: okay \\
    2. \#\#\#Feedback: The translation is not good. The aspect of location of the translation is missing. \\
    Do not comment on the tags(\#\#\#) inside the original or translated text.\\
    \hline
    \end{tabular}
    \end{minipage}

    \vspace{1cm}
    
    \begin{minipage}{\textwidth}
    \begin{tabular}{|p{16cm}|}
    \hline
    \multicolumn{1}{|c|}{\textbf{System Prompt for Agent-3 (GPT-3.5)}} \\
    \hline
        You are an AI assitant who is capable of translating from \{source\_language\} to \{destination\_language\}. You will be given a  from \{source\_language\} and the translation into \{destination\_language\}. \\
        However, there will be some problems in the translation, maybe some incident got missed, some details tweaked or anything like that. You will be given a feedback for that. Closely follow the feedback to improve the translation or insert any missing element in the story or options section. \\
    \hline
    \end{tabular}
    \end{minipage}
    \caption{System Prompts for Agents in the Data Translation Pipeline}
    \label{tab:translation-prompts}
\end{table*}
\section{Cultural Nuance Injection}
\label{appendix:cultural-nuance-injection}
We introduce cultural nuances to the English language dataset before translating them. The reason is twofold. Firstly, models are significantly more effective at generating culturally nuanced content in English. Secondly, the verification process is more straightforward in English. Given that we employ a human-in-the-loop approach to ensure the core narrative of each story and question remains intact, performing this verification in English is more convenient and reliable. We provide the prompting templates used to introduce cultural nuances in Table \ref{tab:cultural-prompts}. 
\\~\\
The use of two separate prompts is necessary for our adaptation process. The adapter prompt takes in the original story, question, and answer choices, along with the adapted story. Its job is to modify the original question and answers to fit the new, adapted story context.
\\~\\
We use this adapter approach because often, a single story is the basis for multiple question-answer pairs. To ensure consistency and allow for valid comparisons between the culturally adapted dataset and the original dataset, it's crucial to keep the adapted story the same for all related question-answer pairs. This method maintains the integrity of our adaptation process and allows us to effectively compare the two datasets.
\begin{table*}
    \centering
    \begin{minipage}{\textwidth}
    \begin{tabular}{|p{16cm}|}
    \hline
    \multicolumn{1}{|c|}{\textbf{System Prompt for Culturally Sensitive Assistant (Llama-3.1)}} \\
    \hline 
        You are a culturally sensitive assistant. Transform the given story to fit the \{language\} culture by adjusting names, settings, and cultural elements while preserving the core narrative.
        Then transform the given Question and the Multiple choices to fit your transformed version of the story, but keep the core idea of the Question and the options constant. Provide a valid JSON as output. Make sure it is a valid JSON format having necessary commas and quotations. \\
    \hline
    \end{tabular}
    \end{minipage}

    \vspace{0.5cm}
    \begin{minipage}{\textwidth}
    \begin{tabular}{|p{16cm}|}
    \hline
    \multicolumn{1}{|c|}{\textbf{System Prompt for Culturally Sensitive Adapter Assistant (Llama-3.1)}} \\
    \hline 
        You are a culturally sensitive assistant. Given a culturally adapted story and the original story, transform the provided original question and multiple-choice
        options to fit the cultural nuances of the story while preserving their core ideas and intent. Do not change answers that contain Knows/Does not Know/None. Do not get creative with the question and answers. Only change in the places absolutely necessary if needed. Ensure that the adapted question and answers are coherent with the culturally nuanced context, while maintaining the original question's core idea and the exact purpose of the options. Then provide a 
        valid JSON as output. Make sure it is valid JSON format having necessary commas and quotations. \\
    \hline
    \end{tabular}
    \end{minipage}
    \caption{System Prompts provided to LLMs for cultural element induction}
    \label{tab:cultural-prompts}
\end{table*}
\begin{table*}
    \centering
    \begin{minipage}{\textwidth}
    \begin{tabular}{c c c c c c c c c c}
    \hline
         & \#S & \#Q & ASL & ASL & ASL & ASL & ASL & ASL & ASL \\
         & & & (En) & (Zh) & (Fr) & (Ru) & (Hi) & (Bn) & (Ar) \\
         \hline
         \textbf{Task View} & 168 & 441 & 95.14 & 67.68 & 98.72 & 89.38 & 99.27 & 63.22 & 88.57 \\ 
         \hline
         Unexpected Outcome Test & 18 & 54 & 
            56.69 &
            37.69 &
            65.56 &
            57.09 &
            62.87 &
            41.93 &
            68.11 \\
         Scalar Implicature Test & 18 & 36 &
            70.67 &
            52.39 &
            81.44 &
            73.94 &
            74.97 &
            51.5 &
            72.14 \\
        Persuasion Story Task & 18 & 18 &
            61.22 &
            39.33 &
            60.61 &
            55.67 &
            58.94 &
            43.89 &
            60.78 \\
        False Belief Task & 18 & 108 &
            70.06 &
            45.54 &
            79.75 &
            65.94 &
            71.19 &
            49.94 &
            67.94 \\
        Ambiguous Story Task & 18 & 36 &
            137.39 &
            105.06 &
            163.22 &
            149.42 &
            176.47 &
            118.83 &
            153.72 \\
        Hinting Task Test & 17 & 20 &
            85.35 &
            57.50 &
            88.30 &
            82.60 &
            88.10 &
            57.05 &
            89.95 \\
       Strange Story Task & 36 & 72 &
            104.33 &
            76.25 &
            115.14 &
            108.03 &
            113.21 &
            78.06 &
            112.72 \\
        Faux-pas Recognition Test & 25 & 97 &
            122.89 &
            92.03 &
            140.12 &
            122.35 &
            144.03 &
            100.26 &
            135.76 \\
        \hline
        \textbf{Ability View} & 214 & 502 &
            91.24 & 65.40 & 102.28 & 92.60 & 101.61 & 70.42 & 97.76 \\
        \hline
        Emotion & 37 & 70 &
            75.14 &
            52.99 &
            84.57 &
            77.26 &
            82.14 &
            56.93 &
            81.51 \\
         Desire & 28 & 32 &
            77.38 &
            53.78 &
            80.28 &
            76.38 &
            80.78 &
            58.38 &
            84.47 \\
        Intention & 51 & 61 &
            116.89 &
            84.26 &
            131.11 &
            120.90 &
            136.13 &
            91.79 &
            126.02 \\
        Knowledge & 29 & 50 &
            81.72 &
            59.34 &
            93.40 &
            84.90 &
            88.90 &
            62.12 &
            84.06 \\
        Belief & 28 & 164 &
            79.20 &
            54.96 &
            91.94 &
            79.37 &
            86.85 &
            59.67 &
            82.52 \\
        Non-literal Communication & 41 & 125 &
            117.14 &
            87.10 &
            132.38 &
            116.82 &
            134.87 &
            93.66 &
            127.96 \\
        \hline            
    \end{tabular}
    \subcaption{Data Statistics for the Dataset with  Cultural Nuance Addition}
    \end{minipage}
    \\~\\
    \vspace{1cm}

    \begin{minipage}{\textwidth}
    \centering
    \begin{tabular}{c c c c c c c c c c}
    \hline
         & \#S & \#Q & ASL & ASL & ASL & ASL & ASL & ASL & ASL \\
         & & & (En) & (Zh) & (Fr) & (Ru) & (Hi) & (Bn) & (Ar) \\
         \hline
         \textbf{Task View} & 168 & 441 & 85.01 & 61.41 & 80.70 & 90.08 & 110.19 & 77.09 & 93.24 \\ 
         \hline
         Unexpected Outcome Test & 18 & 54 & 
            50.96 &
            38.33 &
            60.74 &
            57.20 &
            66.83 &
            45.28 &
            58.11 \\
         Scalar Implicature Test & 18 & 36 &
            64.00 &
            48.78 &
            78.03 &
            68.31 &
            80.56 &
            55.17 &
            71.64 \\
       Persuasion Story Task & 18 & 18 &
        51.00 &
        33.11 &
        60.50 &
        54.89 &
        64.67 &
        47.94 &
        55.00 \\
        False Belief Task & 18 & 108 &
            66.28 &
            44.28 &
            70.44 &
            69.28 &
            88.44 &
            65.44 &
            71.06 \\

        Ambiguous Story Task & 18 & 36 &
            152.33 &
            109.72 &
            82.00 &
            159.50 &
            202.72 &
            137.17 &
            166.00 \\
       Hinting Task Test & 17 & 20 &
        75.45 &
        54.40 &
        89.00 &
        77.80 &
        94.50 &
        66.40 &
        80.05 \\
        
        Strange Story Task & 36 & 72 &
        98.14 &
        69.63 &
        103.25 &
        102.01 &
        120.31 &
        84.39 &
        106.44 \\
        
        Faux-pas Recognition Test & 25 & 97 &
        121.93 &
        93.07 &
        137.59 &
        131.64 &
        163.45 &
        114.93 &
        137.59 \\

        \hline
        \textbf{Ability View} & 214 & 502 & 87.11 & 63.16 & 88.04 & 92.17 & 113.00 & 78.63 & 95.69 \\
        \hline
        Emotion & 37 & 70 &
            68.39 &
            51.11 &
            74.77 &
            74.80 &
            87.97 &
            60.53 &
            76.66 \\
         Desire & 28 & 32 &
            69.00 &
            47.31 &
            80.34 &
            71.13 &
            88.41 &
            62.59 &
            75.50 \\
        Intention & 51 & 61 &
            116.46 &
            83.18 &
            108.00 &
            120.74 &
            150.52 &
            102.75 &
            125.38 \\

        Knowledge & 29 & 50 &
            76.34 &
            56.98 &
            90.86 &
            81.30 &
            97.86 &
            66.50 &
            84.56 \\
        Belief & 28 & 164 &
            77.59 &
            53.84 &
            74.27 &
            82.18 &
            102.30 &
            73.18 &
            84.03 \\
        Non-literal Communication & 41 & 125 &
            114.91 &
            86.56 &
            100.00 &
            122.87 &
            150.92 &
            106.22 &
            128.00 \\
        \hline            
    \end{tabular}
    \subcaption{Data Statistics for the Direct Translated Dataset}
    \end{minipage}
    \caption{The tables present the dataset statistics for both the Directly Translated Dataset and the dataset with Cultural Elements Introduced. \#S: The Number of stories, \#Q: The Number of Questions, ASL: Average Story Length.}
    \label{tab:generic-cultural-stats}
\end{table*}

\begin{figure*}[h]
    \centering
    \begin{subfigure}[b]{.9\textwidth}
    \centering
    \includegraphics[width=\textwidth]{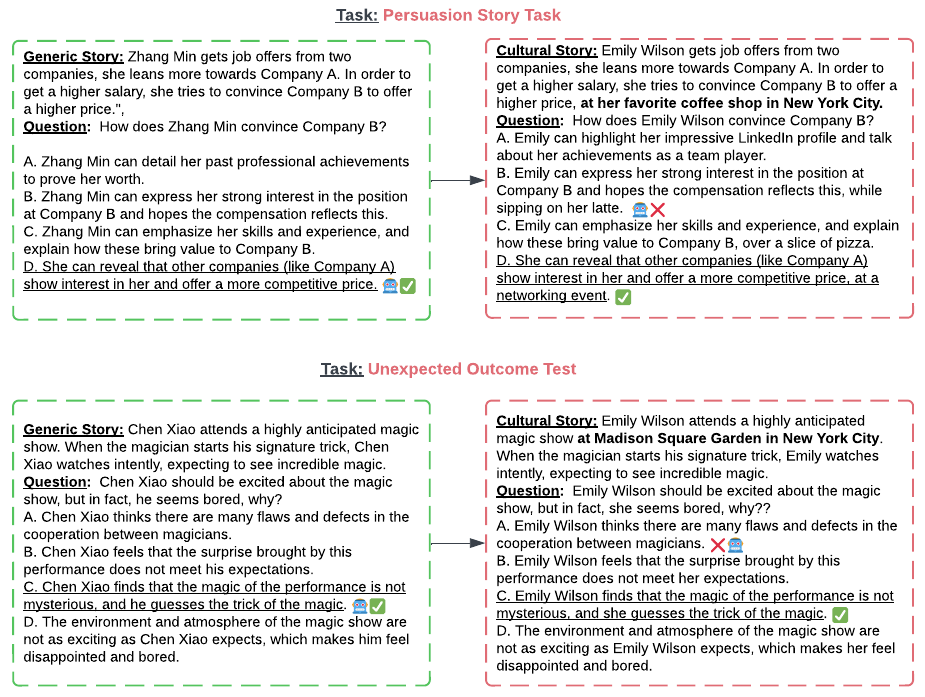}
    \caption{English Cultural nuances Added}
    \label{fig:english-incoherence}
    \end{subfigure}
    \hfill
    \begin{subfigure}[b]{.9\textwidth}
        \centering
        \includegraphics[width=\textwidth]{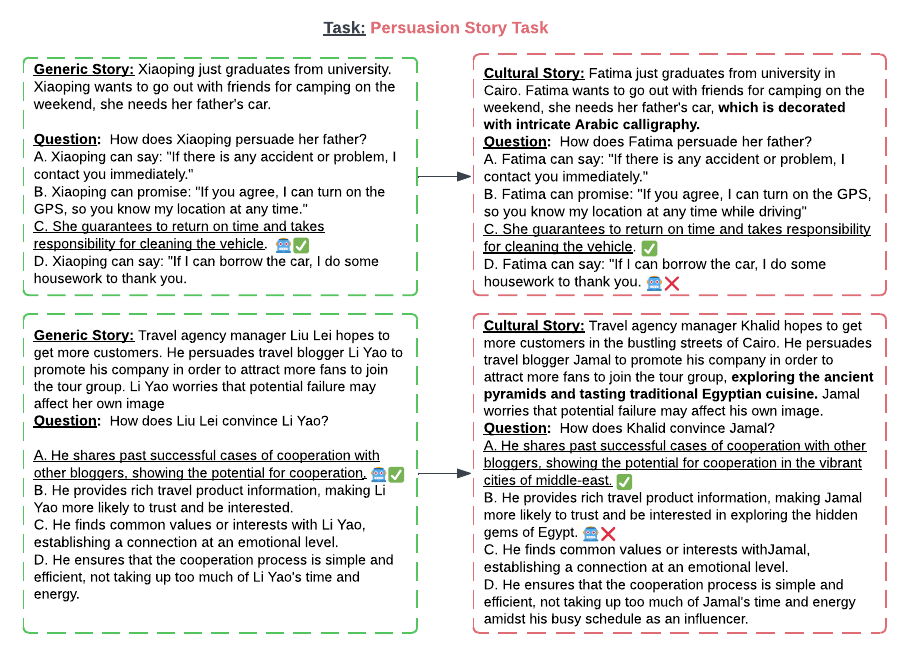}
        \caption{Arabic Cultural nuances Added}
        \label{fig:arabic-incoherence}
    \end{subfigure}
    \caption{Comparison of Generic and Culturally adapted Tasks: illustrating model tendency to change answer choices when cultural nuances are added}
    \label{fig:cultural-incoherence}
\end{figure*}

\section{Generated Results by LLMs in Detail}

We provide all our results in details in this section. We provide the raw response for tasks and abilities for each model for both direct translated dataset and culturally nuanced dataset from Tables \ref{tab:claude_sonnet_data_table} to \ref{tab:llama2_data_table}.

\begin{figure*}[htbp]
    \centering
    \begin{subfigure}[b]{\textwidth}
        \centering
        \includegraphics[width=\textwidth]{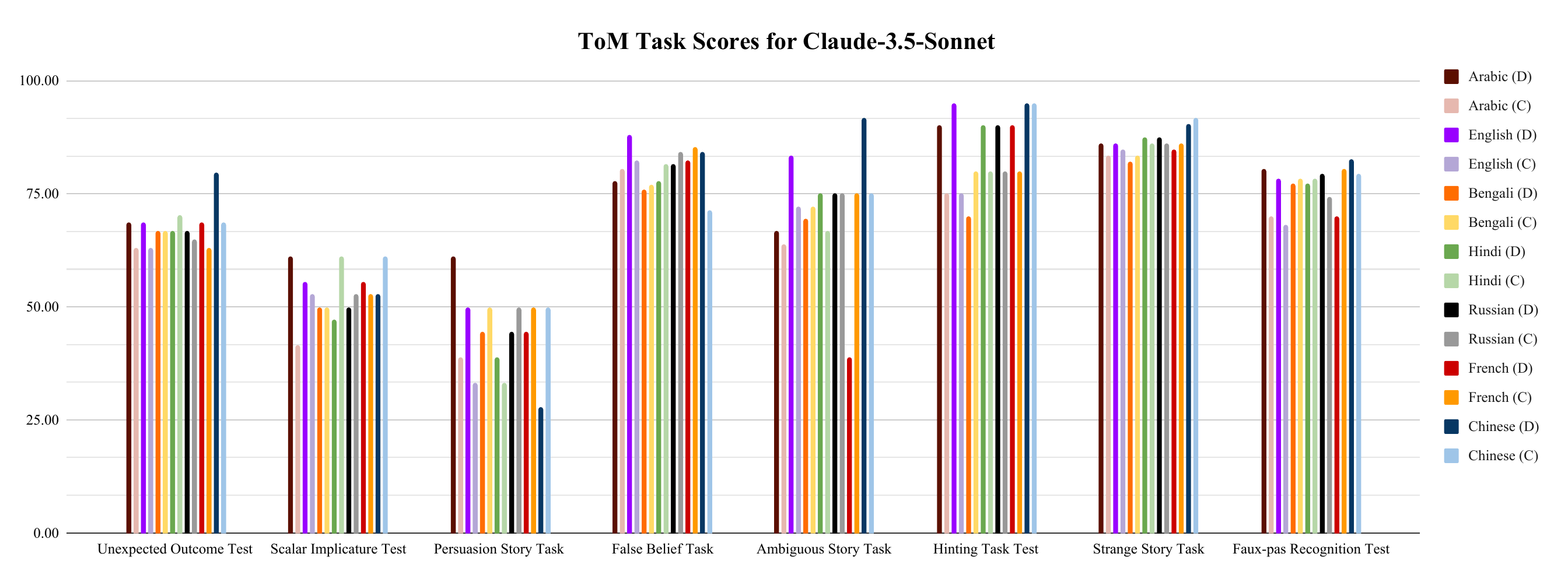}
        \caption{ToM Task Scores for Claude-3.5-Sonnet}
        \label{fig:tom_task_Claude-3.5}
    \end{subfigure}
    \hfill
    \begin{subfigure}[b]{\textwidth}
        \centering
        \includegraphics[width=\textwidth]{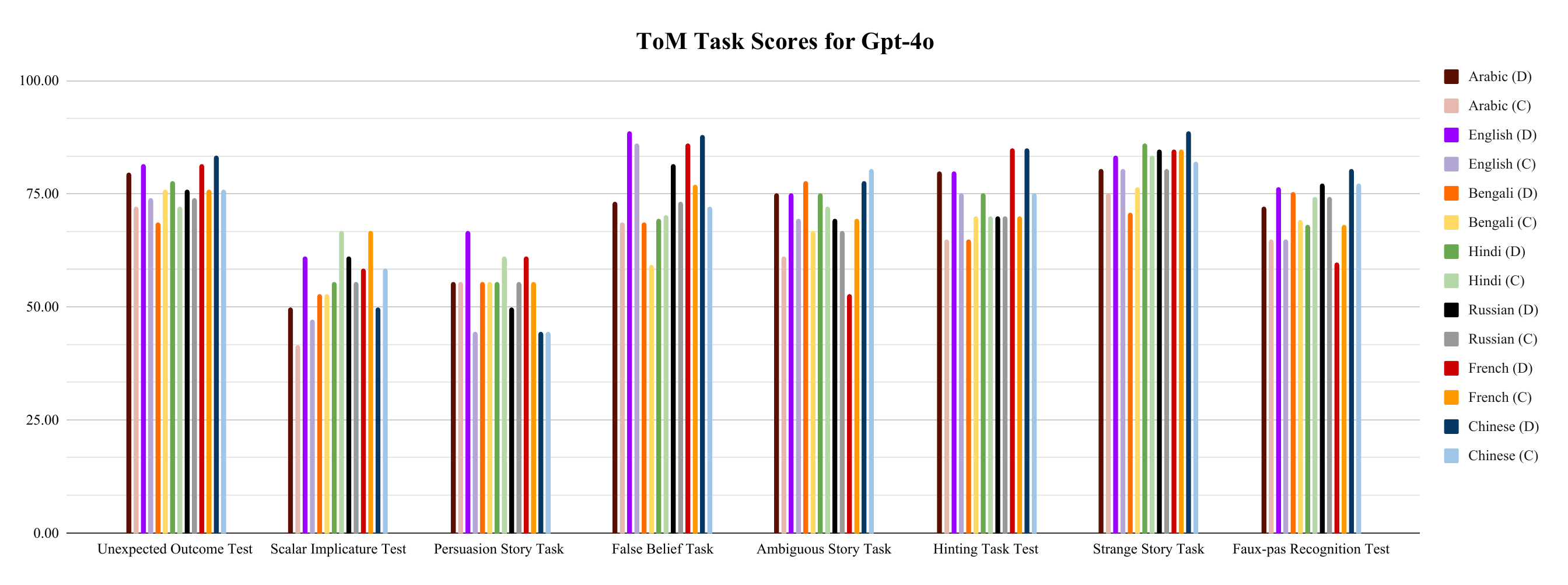}
        \caption{ToM Task Scores for Gpt-4o}
        \label{fig:tom_task_gpt4}
    \end{subfigure}
    \hfill
    \begin{subfigure}[b]{\textwidth}
        \centering
        \includegraphics[width=\textwidth]{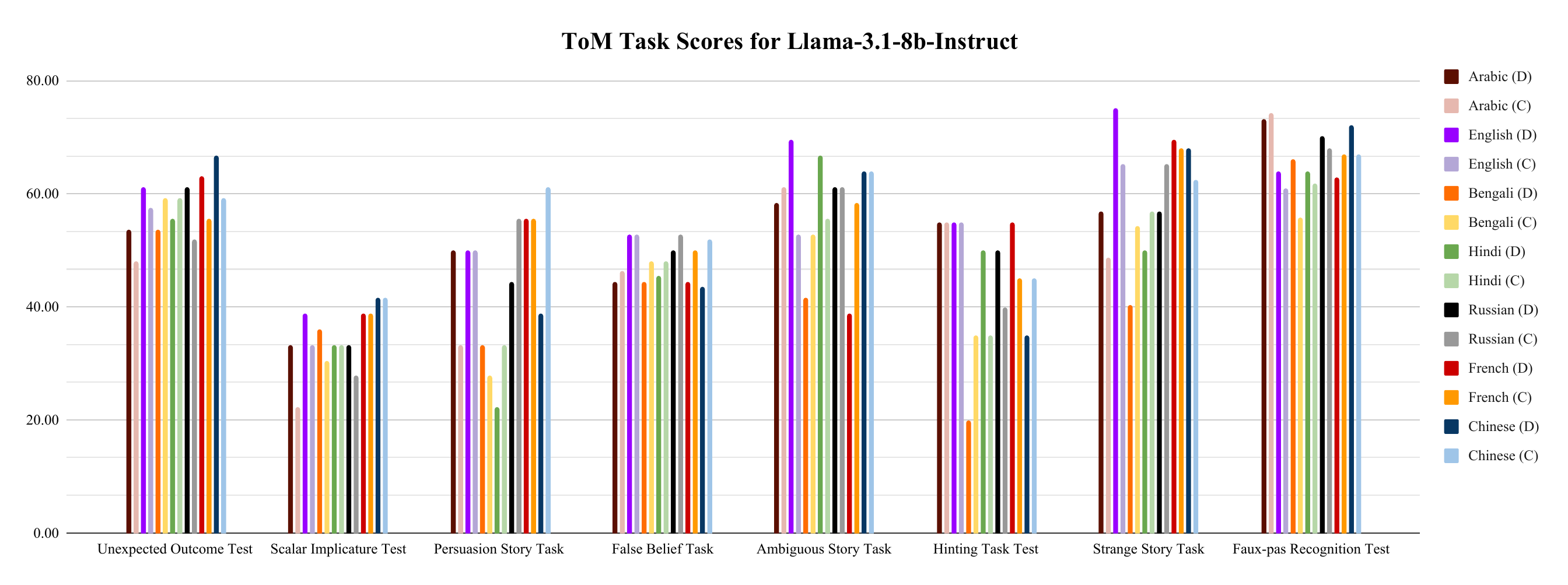}
        \caption{ToM Task Scores for Llama-3.1-8b-Instruct}
        \label{fig:tom_task_llama3}
    \end{subfigure}
    
    \caption{ToM Task Scores for Three Stronger LLMs}
    \label{fig:mainfig1}
\end{figure*}

\begin{figure*}[htbp]
    \centering
    \begin{subfigure}[b]{\textwidth}
        \centering
        \includegraphics[width=\textwidth]{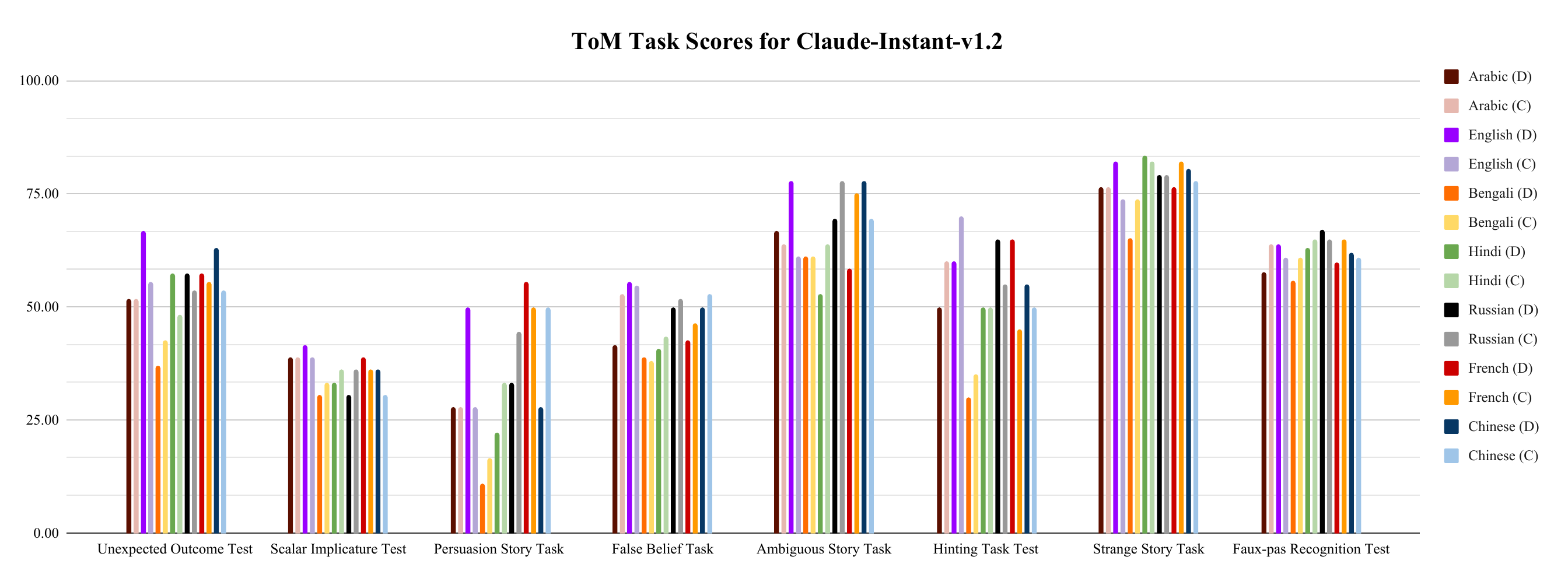}
        \caption{ToM Task Scores for Claude-Instant-v1.2}
        \label{fig:tom_task_claude_inst}
    \end{subfigure}
    \hfill
    \begin{subfigure}[b]{\textwidth}
        \centering
        \includegraphics[width=\textwidth]{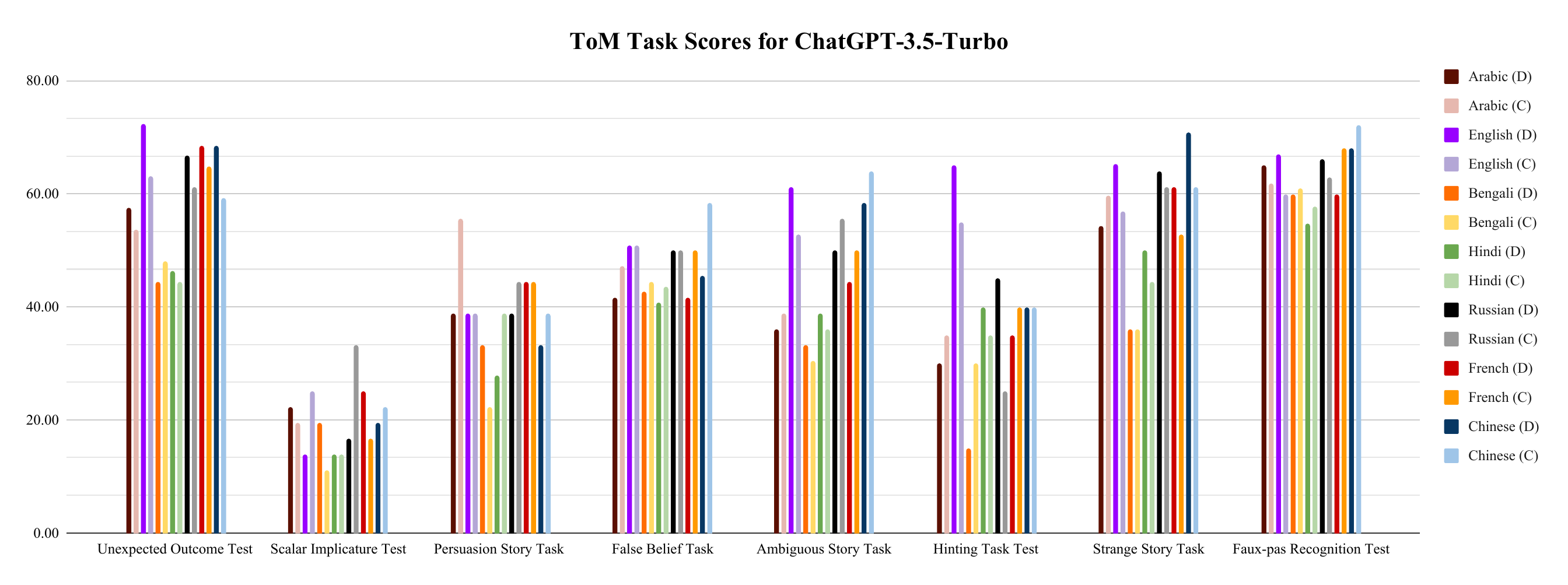}
        \caption{ToM Task Scores for ChatGPT-3.5-Turbo}
        \label{fig:tom_task_chatgpt3}
    \end{subfigure}
    \hfill
    \begin{subfigure}[b]{\textwidth}
        \centering
        \includegraphics[width=\textwidth]{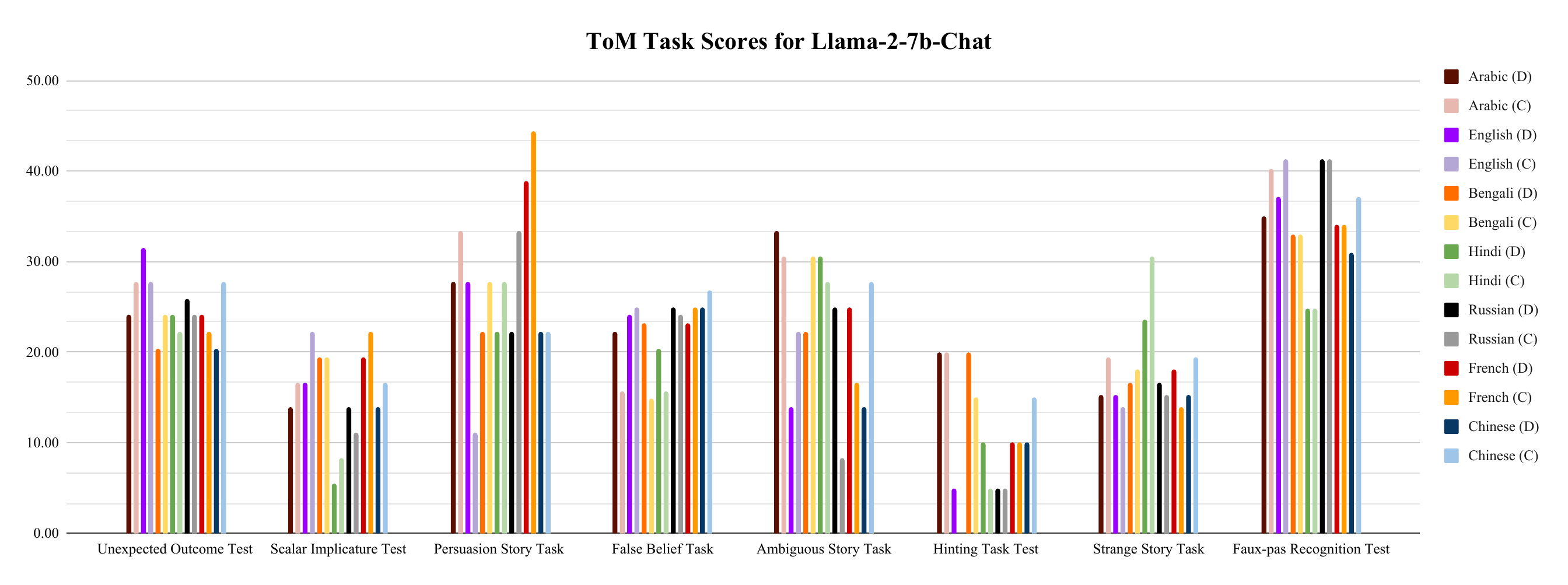}
        \caption{ToM Task Scores for Llama-2-7b-Chat}
        \label{fig:tom_task_llama2}
    \end{subfigure}
    
    \caption{ToM Task Scores for Three Weaker LLMs}
    \label{fig:mainfig2}
\end{figure*}

\begin{figure*}[htbp]
    \centering
    \begin{subfigure}[b]{\textwidth}
        \centering
        \includegraphics[width=\textwidth]{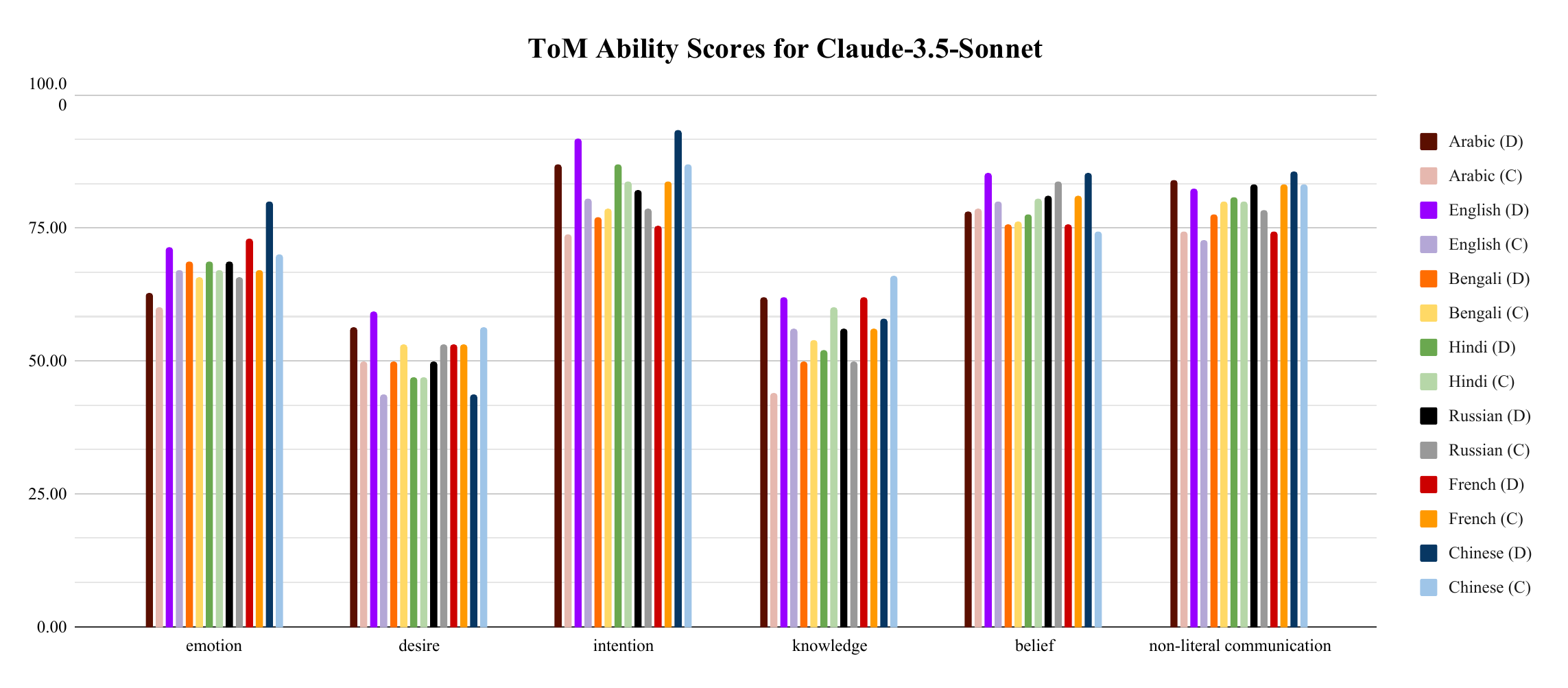}
        \caption{ToM Ability Scores for Claude-3.5-Sonnet}
        \label{fig:tom_ability_claude3}
    \end{subfigure}
    \hfill
    \begin{subfigure}[b]{\textwidth}
        \centering
        \includegraphics[width=\textwidth]{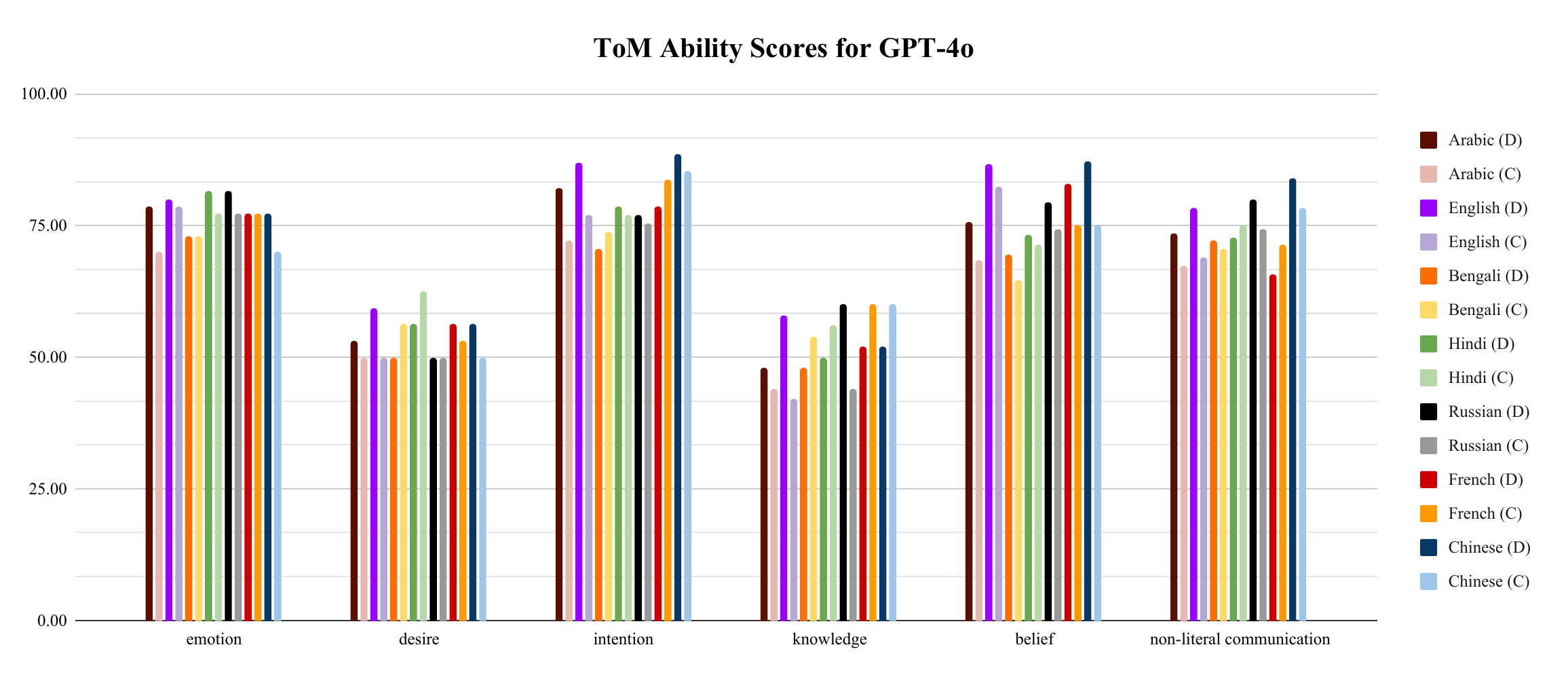}
        \caption{ToM Ability Scores for Gpt-4o}
        \label{fig:tom_ability_gpt4}
    \end{subfigure}
    \hfill
    \begin{subfigure}[b]{\textwidth}
        \centering
        \includegraphics[width=\textwidth]{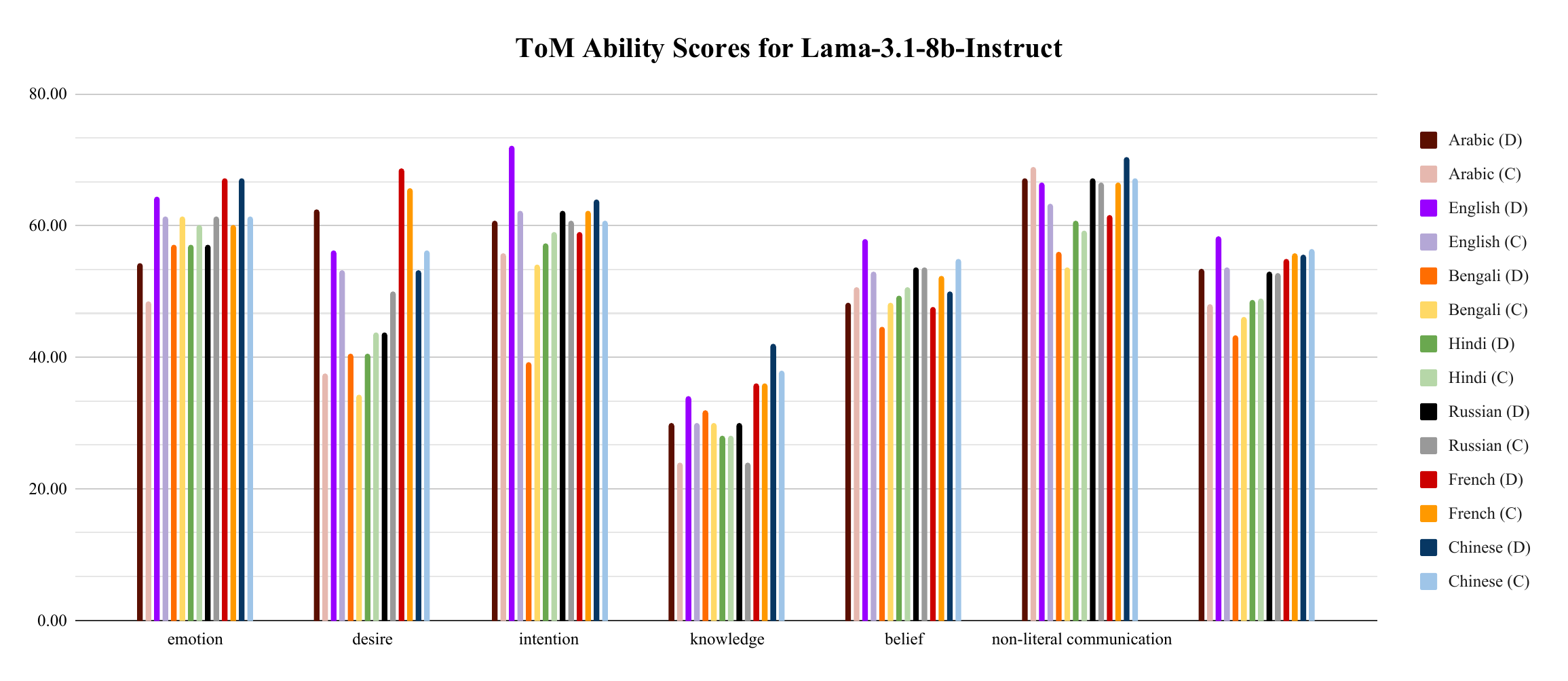}
        \caption{ToM Ability Scores for Lama-3.1-8b-Instruct}
        \label{fig:tom_ability_llama3}
    \end{subfigure}
    
    \caption{ToM Ability Score for Three Stronger LLMs}
    \label{fig:mainfig3}
\end{figure*}

\begin{figure*}[htbp]
    \centering
    \begin{subfigure}[b]{\textwidth}
        \centering
        \includegraphics[width=\textwidth]{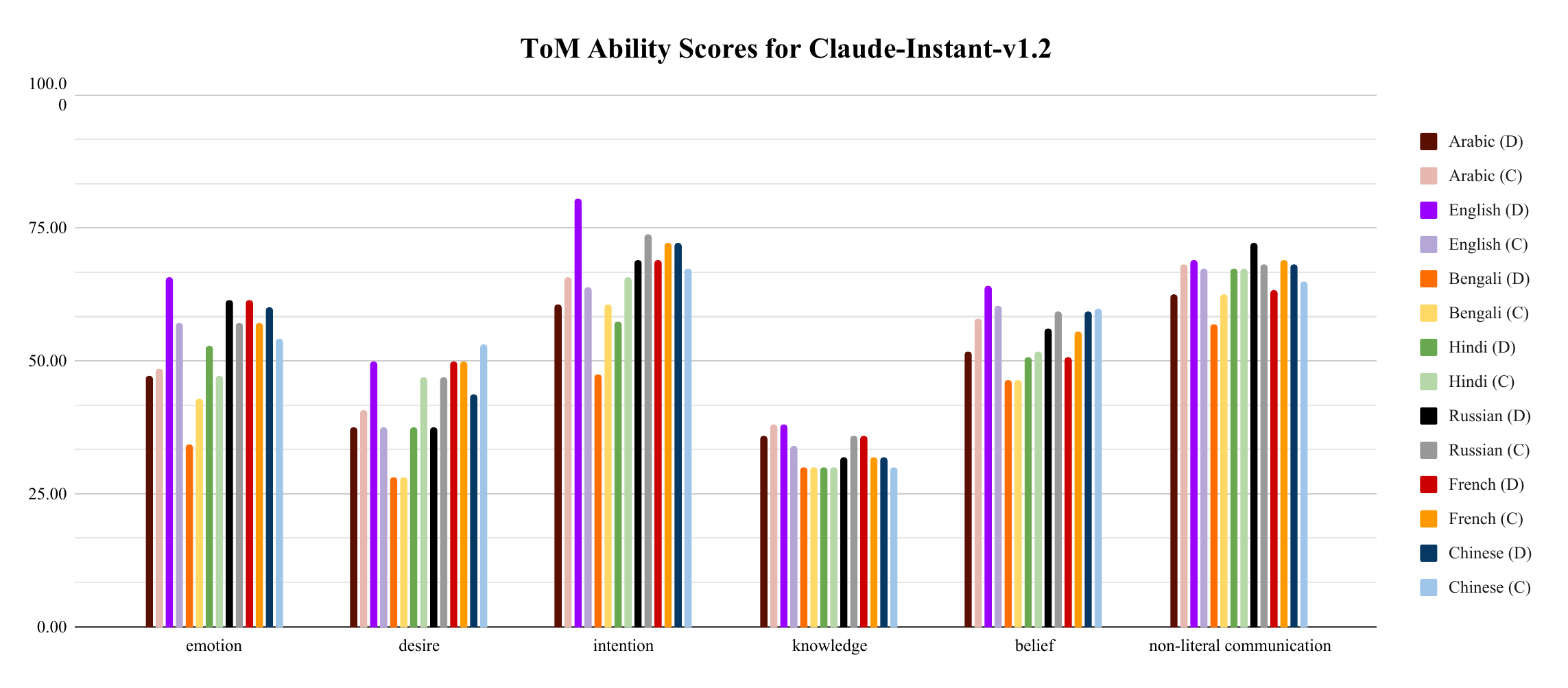}
        \caption{ToM Ability Scores for Claude-Instant-v1.2}
        \label{fig:tom_ability_claude_inst}
    \end{subfigure}
    \hfill
    \begin{subfigure}[b]{\textwidth}
        \centering
        \includegraphics[width=\textwidth]{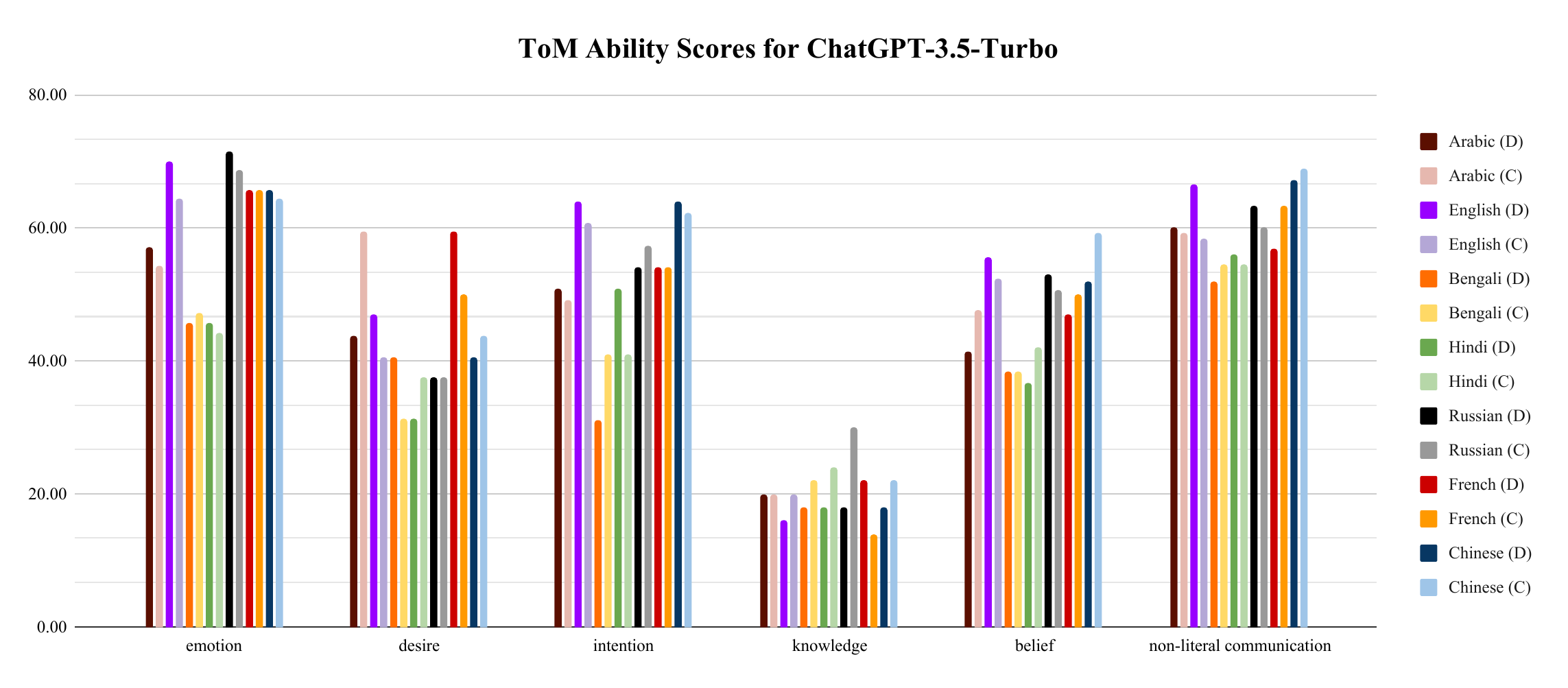}
        \caption{ToM Ability Scores for ChatGPT-3.5-Turbo}
        \label{fig:tom_ability_gpt3}
    \end{subfigure}
    \hfill
    \begin{subfigure}[b]{\textwidth}
        \centering
        \includegraphics[width=\textwidth]{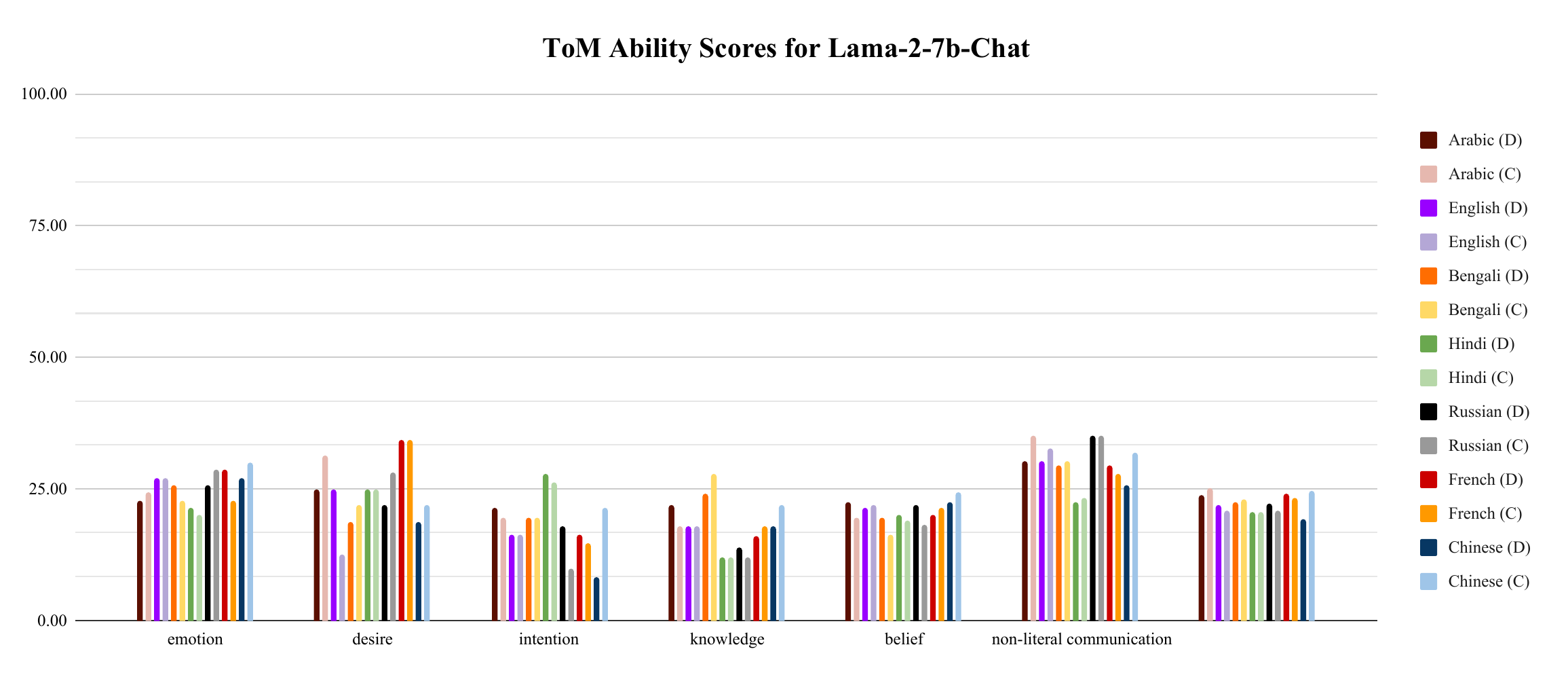}
        \caption{ToM Ability Scores for Llama-2-7b-Chat}
        \label{fig:tom_ability_llama2}
    \end{subfigure}
    
    \caption{Tom Ability Scores for Three Weaker LLMs}
    \label{fig:mainfig4}
\end{figure*}

In Figure \ref{fig:mainfig1} and \ref{fig:mainfig2} we present a comparative study of the ToM task scores for different languages with results from direct translations and culturally nuanced translations for individual LLMs. Similar data visualization is shown in \ref{fig:mainfig3} and \ref{fig:mainfig4}, with the only difference being those scores are for ToM abilities. 

\label{appendix:result-details}









\section{Cultural Incoherence in Theory of Mind}
\label{appendix:incoherence-tests}

Our analysis reveals a notable decline in model accuracy when specific cultural nuances are introduced into the task. Ideally, models should demonstrate the ability to disregard irrelevant information within a question. However, in scenarios involving cultural adaptations, we observe a tendency for models to select incorrect answer choices. This stands in contrast to their performance on direct translations, where models successfully identify the correct answers.
\\~\\
This phenomenon suggests a potential vulnerability in the models' ability to navigate culturally nuanced information while maintaining task-relevant focus. It raises important questions about the robustness of language models in handling diverse cultural contexts and their capacity to locate relevant information across culturally nuanced ToM tasks. Figure \ref{fig:cultural-incoherence} illustrates an example.

\begin{table*}[h!]
    \centering
    \small
    \textbf{Abbreviations:} \\
    UOT: Unexpected Outcome Test, SIT: Scalar Implicature Test, PST: Persuasion Story Task, \\
    FBT: False Belief Task, AST: Ambiguous Story Task, HTT: Hinting Task Test, SST: Strange Story Task, \\
    FRT: Faux-pas Recognition Test, EMO: Emotion, DES: Desire, INT: Intention, \\
    KNO: Knowledge, BEL: Belief, NLC: Non-literal Communication \\
    \vspace{0.3cm}

    \begin{tabular}{|l|l|c|c|c|c|c|c|c|c|c|c|c|c|c|c|}
        \hline
        \textbf{Lang} & & \textbf{UOT} & \textbf{SIT} & \textbf{PST} & \textbf{FBT} & \textbf{AST} & \textbf{HTT} & \textbf{SST} & \textbf{FRT} & \textbf{EMO} & \textbf{DES} & \textbf{INT} & \textbf{KNO} & \textbf{BEL} & \textbf{NLC} \\
        \hline
        \multirow{2}{*}{\textbf{Arabic}} & D & 68.5 & 61.1 & 61.1 & 77.8 & 66.7 & 90.0 & 86.1 & 80.4 & 62.9 & 56.3 & 86.9 & 62.0 & 78.1 & 84.0 \\
        & C & 63.0 & 41.7 & 38.9 & 80.6 & 63.9 & 75.0 & 83.3 & 70.1 & 60.0 & 50.0 & 73.8 & 44.0 & 78.7 & 74.4 \\
        \hline
        \multirow{2}{*}{\textbf{English}} & D & 68.5 & 55.6 & 50.0 & 88.0 & 83.3 & 95.0 & 86.1 & 78.4 & 71.4 & 59.4 & 91.8 & 62.0 & 85.4 & 82.4 \\
        & C & 63.0 & 52.8 & 33.3 & 82.4 & 72.2 & 75.0 & 84.7 & 68.0 & 67.1 & 43.8 & 80.3 & 56.0 & 79.9 & 72.8 \\
        \hline
        \multirow{2}{*}{\textbf{Bengali}} & D & 66.7 & 50.0 & 44.4 & 75.9 & 69.4 & 70.0 & 81.9 & 77.3 & 68.6 & 50.0 & 77.1 & 50.0 & 75.6 & 77.6 \\
        & C & 66.7 & 50.0 & 50.0 & 76.9 & 72.2 & 80.0 & 83.3 & 78.4 & 65.7 & 53.1 & 78.7 & 54.0 & 76.2 & 80.0 \\
        \hline
        \multirow{2}{*}{\textbf{Hindi}} & D & 66.7 & 47.2 & 38.9 & 77.8 & 75.0 & 90.0 & 87.5 & 77.3 & 68.6 & 46.9 & 86.9 & 52.0 & 77.4 & 80.8 \\
        & C & 70.4 & 61.1 & 33.3 & 81.5 & 66.7 & 80.0 & 86.1 & 78.4 & 67.1 & 46.9 & 83.6 & 60.0 & 80.5 & 80.0 \\
        \hline
        \multirow{2}{*}{\textbf{Russian}} & D & 66.7 & 50.0 & 44.4 & 81.5 & 75.0 & 90.0 & 87.5 & 79.4 & 68.6 & 50.0 & 82.0 & 56.0 & 81.1 & 83.2 \\
        & C & 64.8 & 52.8 & 50.0 & 84.3 & 75.0 & 80.0 & 86.1 & 74.2 & 65.7 & 53.1 & 78.7 & 50.0 & 83.5 & 78.4 \\
        \hline
        \multirow{2}{*}{\textbf{French}} & D & 68.5 & 55.6 & 44.4 & 82.4 & 38.9 & 90.0 & 84.7 & 70.1 & 72.9 & 53.1 & 75.4 & 62.0 & 75.6 & 74.4 \\
        & C & 63.0 & 52.8 & 50.0 & 85.2 & 75.0 & 80.0 & 86.1 & 80.4 & 67.1 & 53.1 & 83.6 & 56.0 & 81.1 & 83.2 \\
        \hline
        \multirow{2}{*}{\textbf{Chinese}} & D & 79.6 & 52.8 & 27.8 & 84.3 & 91.7 & 95.0 & 90.3 & 82.5 & 80.0 & 43.8 & 93.4 & 58.0 & 85.4 & 85.6 \\
        & C & 68.5 & 61.1 & 50.0 & 71.3 & 75.0 & 95.0 & 91.7 & 79.4 & 70.0 & 56.3 & 86.9 & 66.0 & 74.4 & 83.2 \\
        \hline
    \end{tabular}
    \caption{Performance of Various Tasks and Abilities across Languages and Cultural Contexts (D: Direct/, C: Cultural) for \textbf{Claude-3.5-Sonnet}}
    \label{tab:claude_sonnet_data_table}
\end{table*}

\begin{table*}[h!]
    \centering
    \small

    \begin{tabular}{|l|l|c|c|c|c|c|c|c|c|c|c|c|c|c|c|}
        \hline
        \textbf{Lang} & & \textbf{UOT} & \textbf{SIT} & \textbf{PST} & \textbf{FBT} & \textbf{AST} & \textbf{HTT} & \textbf{SST} & \textbf{FRT} & \textbf{EMO} & \textbf{DES} & \textbf{INT} & \textbf{KNO} & \textbf{BEL} & \textbf{NLC} \\
        \hline
            \multirow{2}{*}{\textbf{Arabic}} & D & 51.9 & 38.9 & 27.8 & 41.7 & 66.7 & 50.0 & 76.4 & 57.7 & 47.1 & 37.5 & 60.7 & 36.0 & 51.8 & 62.4 \\
        & C & 51.9 & 38.9 & 27.8 & 52.8 & 63.9 & 60.0 & 76.4 & 63.9 & 48.6 & 40.6 & 65.6 & 38.0 & 57.9 & 68.0 \\
        \hline
        \multirow{2}{*}{\textbf{English }} & D & 66.7 & 41.7 & 50.0 & 55.6 & 77.8 & 60.0 & 81.9 & 63.9 & 65.7 & 50.0 & 80.3 & 38.0 & 64.0 & 68.8 \\
        & C & 55.6 & 38.9 & 27.8 & 54.6 & 61.1 & 70.0 & 73.6 & 60.8 & 57.1 & 37.5 & 63.9 & 34.0 & 60.4 & 67.2 \\
        \hline
        \multirow{2}{*}{\textbf{Bengali }} & D & 37.0 & 30.6 & 11.1 & 38.9 & 61.1 & 30.0 & 65.3 & 55.7 & 34.3 & 28.1 & 47.5 & 30.0 & 46.3 & 56.8 \\
        & C & 42.6 & 33.3 & 16.7 & 38.0 & 61.1 & 35.0 & 73.6 & 60.8 & 42.9 & 28.1 & 60.7 & 30.0 & 46.3 & 62.4 \\
        \hline
        \multirow{2}{*}{\textbf{Hindi }} & D & 57.4 & 33.3 & 22.2 & 40.7 & 52.8 & 50.0 & 83.3 & 62.9 & 52.9 & 37.5 & 57.4 & 30.0 & 50.6 & 67.2 \\
        & C & 48.1 & 36.1 & 33.3 & 43.5 & 63.9 & 50.0 & 81.9 & 64.9 & 47.1 & 46.9 & 65.6 & 30.0 & 51.8 & 67.2 \\
        \hline
        \multirow{2}{*}{\textbf{Russian }} & D & 57.4 & 30.6 & 33.3 & 50.0 & 69.4 & 65.0 & 79.2 & 67.0 & 61.4 & 37.5 & 68.9 & 32.0 & 56.1 & 72.0 \\
        & C & 53.7 & 36.1 & 44.4 & 51.9 & 77.8 & 55.0 & 79.2 & 64.9 & 57.1 & 46.9 & 73.8 & 36.0 & 59.1 & 68.0 \\
        \hline
        \multirow{2}{*}{\textbf{French }} & D & 57.4 & 38.9 & 55.6 & 42.6 & 58.3 & 65.0 & 76.4 & 59.8 & 61.4 & 50.0 & 68.9 & 36.0 & 50.6 & 63.2 \\
        & C & 55.6 & 36.1 & 50.0 & 46.3 & 75.0 & 45.0 & 81.9 & 64.9 & 57.1 & 50.0 & 72.1 & 32.0 & 55.5 & 68.8 \\
        \hline
        \multirow{2}{*}{\textbf{Chinese }} & D & 63.0 & 36.1 & 27.8 & 50.0 & 77.8 & 55.0 & 80.6 & 61.9 & 60.0 & 43.8 & 72.1 & 32.0 & 59.1 & 68.0 \\
        & C & 53.7 & 30.6 & 50.0 & 52.8 & 69.4 & 50.0 & 77.8 & 60.8 & 54.3 & 53.1 & 67.2 & 30.0 & 59.8 & 64.8 \\
        \hline

    \hline
    \end{tabular}
    \caption{Performance of Various Tasks and Abilities across Languages and Cultural Contexts (D: Direct/, C: Cultural) for \textbf{Claude-Instant-1.2}}
    \label{tab:claude_inst_data_table}
\end{table*}

\begin{table*}[h!]
    \centering
    \small
    \begin{tabular}{|l|l|c|c|c|c|c|c|c|c|c|c|c|c|c|c|}
        \hline
        \textbf{Lang} & & \textbf{UOT} & \textbf{SIT} & \textbf{PST} & \textbf{FBT} & \textbf{AST} & \textbf{HTT} & \textbf{SST} & \textbf{FRT} & \textbf{EMO} & \textbf{DES} & \textbf{INT} & \textbf{KNO} & \textbf{BEL} & \textbf{NLC} \\
        \hline
            \multirow{2}{*}{\textbf{Arabic}} & D & 79.6 & 50.0 & 55.6 & 73.1 & 75.0 & 80.0 & 80.6 & 72.2 & 78.6 & 53.1 & 82.0 & 48.0 & 75.6 & 73.6 \\
        & C & 72.2 & 41.7 & 55.6 & 68.5 & 61.1 & 65.0 & 75.0 & 64.9 & 70.0 & 50.0 & 72.1 & 44.0 & 68.3 & 67.2 \\
        \hline
        \multirow{2}{*}{\textbf{English}} & D & 81.5 & 61.1 & 66.7 & 88.9 & 75.0 & 80.0 & 83.3 & 76.3 & 80.0 & 59.4 & 86.9 & 58.0 & 86.6 & 78.4 \\
        & C & 74.1 & 47.2 & 44.4 & 86.1 & 69.4 & 75.0 & 80.6 & 64.9 & 78.6 & 50.0 & 77.0 & 42.0 & 82.3 & 68.8 \\
        \hline
        \multirow{2}{*}{\textbf{Bengali}} & D & 68.5 & 52.8 & 55.6 & 68.5 & 77.8 & 65.0 & 70.8 & 75.3 & 72.9 & 50.0 & 70.5 & 48.0 & 69.5 & 72.0 \\
        & C & 75.9 & 52.8 & 55.6 & 59.3 & 66.7 & 70.0 & 76.4 & 69.1 & 72.9 & 56.2 & 73.8 & 54.0 & 64.6 & 70.4 \\
        \hline
        \multirow{2}{*}{\textbf{Hindi}} & D & 77.8 & 55.6 & 55.6 & 69.4 & 75.0 & 75.0 & 86.1 & 68.0 & 81.4 & 56.2 & 78.7 & 50.0 & 73.2 & 72.8 \\
        & C & 72.2 & 66.7 & 61.1 & 70.4 & 72.2 & 70.0 & 83.3 & 74.2 & 77.1 & 62.5 & 77.0 & 56.0 & 71.3 & 75.2 \\
        \hline
        \multirow{2}{*}{\textbf{Russian}} & D & 75.9 & 61.1 & 50.0 & 81.5 & 69.4 & 70.0 & 84.7 & 77.3 & 81.4 & 50.0 & 77.0 & 60.0 & 79.3 & 80.0 \\
        & C & 74.1 & 55.6 & 55.6 & 73.1 & 66.7 & 70.0 & 80.6 & 74.2 & 77.1 & 50.0 & 75.4 & 44.0 & 74.4 & 74.4 \\
        \hline
        \multirow{2}{*}{\textbf{French}} & D & 81.5 & 58.3 & 61.1 & 86.1 & 52.8 & 85.0 & 84.7 & 59.8 & 77.1 & 56.2 & 78.7 & 52.0 & 82.9 & 65.6 \\
        & C & 75.9 & 66.7 & 55.6 & 76.9 & 69.4 & 70.0 & 84.7 & 68.0 & 77.1 & 53.1 & 83.6 & 60.0 & 75.0 & 71.2 \\
        \hline
        \multirow{2}{*}{\textbf{Chinese}} & D & 83.3 & 50.0 & 44.4 & 88.0 & 77.8 & 85.0 & 88.9 & 80.4 & 77.1 & 56.2 & 88.5 & 52.0 & 87.2 & 84.0 \\
        & C & 75.9 & 58.3 & 44.4 & 72.2 & 80.6 & 75.0 & 81.9 & 77.3 & 70.0 & 50.0 & 85.2 & 60.0 & 75.0 & 78.4 \\
        \hline

    \hline
    \end{tabular}
    \caption{Performance of Various Tasks and Abilities across Languages and Cultural Contexts (D: Direct/, C: Cultural) for \textbf{GPT-4o}}
    \label{tab:gpt_4_data_table}
\end{table*}

\begin{table*}[h!]
    \centering
    \small
    \textbf{Abbreviations:} \\
    UOT: Unexpected Outcome Test, SIT: Scalar Implicature Test, PST: Persuasion Story Task, \\
    FBT: False Belief Task, AST: Ambiguous Story Task, HTT: Hinting Task Test, SST: Strange Story Task, \\
    FRT: Faux-pas Recognition Test, EMO: Emotion, DES: Desire, INT: Intention, \\
    KNO: Knowledge, BEL: Belief, NLC: Non-literal Communication \\
    \vspace{0.3cm}

    \begin{tabular}{|l|l|c|c|c|c|c|c|c|c|c|c|c|c|c|c|}
        \hline
        \textbf{Lang} & & \textbf{UOT} & \textbf{SIT} & \textbf{PST} & \textbf{FBT} & \textbf{AST} & \textbf{HTT} & \textbf{SST} & \textbf{FRT} & \textbf{EMO} & \textbf{DES} & \textbf{INT} & \textbf{KNO} & \textbf{BEL} & \textbf{NLC} \\
        \hline
            \multirow{2}{*}{\textbf{Arabic}} & D & 57.4 & 22.2 & 38.9 & 41.7 & 36.1 & 30.0 & 54.2 & 64.9 & 57.1 & 43.8 & 50.8 & 20.0 & 41.5 & 60.0 \\
        & C & 53.7 & 19.4 & 55.6 & 47.2 & 38.9 & 35.0 & 59.7 & 61.9 & 54.3 & 59.4 & 49.2 & 20.0 & 47.6 & 59.2 \\
        \hline
        \multirow{2}{*}{\textbf{English}} & D & 72.2 & 13.9 & 38.9 & 50.9 & 61.1 & 65.0 & 65.3 & 67.0 & 70.0 & 46.9 & 63.9 & 16.0 & 55.5 & 66.4 \\
        & C & 63.0 & 25.0 & 38.9 & 50.9 & 52.8 & 55.0 & 56.9 & 59.8 & 64.3 & 40.6 & 60.7 & 20.0 & 52.4 & 58.4 \\
        \hline
        \multirow{2}{*}{\textbf{Bengali}} & D & 44.4 & 19.4 & 33.3 & 42.6 & 33.3 & 15.0 & 36.1 & 59.8 & 45.7 & 40.6 & 31.1 & 18.0 & 38.4 & 52.0 \\
        & C & 48.1 & 11.1 & 22.2 & 44.4 & 30.6 & 30.0 & 36.1 & 60.8 & 47.1 & 31.2 & 41.0 & 22.0 & 38.4 & 54.4 \\
        \hline
        \multirow{2}{*}{\textbf{Hindi}} & D & 46.3 & 13.9 & 27.8 & 40.7 & 38.9 & 40.0 & 50.0 & 54.6 & 45.7 & 31.2 & 50.8 & 18.0 & 36.6 & 56.0 \\
        & C & 44.4 & 13.9 & 38.9 & 43.5 & 36.1 & 35.0 & 44.4 & 57.7 & 44.3 & 37.5 & 41.0 & 24.0 & 42.1 & 54.4 \\
        \hline
        \multirow{2}{*}{\textbf{Russian}} & D & 66.7 & 16.7 & 38.9 & 50.0 & 50.0 & 45.0 & 63.9 & 66.0 & 71.4 & 37.5 & 54.1 & 18.0 & 53.0 & 63.2 \\
        & C & 61.1 & 33.3 & 44.4 & 50.0 & 55.6 & 25.0 & 61.1 & 62.9 & 68.6 & 37.5 & 57.4 & 30.0 & 50.6 & 60.0 \\
        \hline
        \multirow{2}{*}{\textbf{French}} & D & 68.5 & 25.0 & 44.4 & 41.7 & 44.4 & 35.0 & 61.1 & 59.8 & 65.7 & 59.4 & 54.1 & 22.0 & 47.0 & 56.8 \\
        & C & 64.8 & 16.7 & 44.4 & 50.0 & 50.0 & 40.0 & 52.8 & 68.0 & 65.7 & 50.0 & 54.1 & 14.0 & 50.0 & 63.2 \\
        \hline
        \multirow{2}{*}{\textbf{Chinese}} & D & 68.5 & 19.4 & 33.3 & 45.4 & 58.3 & 40.0 & 70.8 & 68.0 & 65.7 & 40.6 & 63.9 & 18.0 & 51.8 & 67.2 \\
        & C & 59.3 & 22.2 & 38.9 & 58.3 & 63.9 & 40.0 & 61.1 & 72.2 & 64.3 & 43.8 & 62.3 & 22.0 & 59.1 & 68.8 \\
        \hline

    \hline
    \end{tabular}
    \caption{Performance of Various Tasks and Abilities across Languages and Cultural Contexts (D: Direct/, C: Cultural) for \textbf{GPT-3.5-Turbo}}
    \label{tab:gpt_3_5_data_table}
\end{table*}

\begin{table*}[h!]
    \centering
    \small

    \begin{tabular}{|l|l|c|c|c|c|c|c|c|c|c|c|c|c|c|c|}
        \hline
        \textbf{Lang} & & \textbf{UOT} & \textbf{SIT} & \textbf{PST} & \textbf{FBT} & \textbf{AST} & \textbf{HTT} & \textbf{SST} & \textbf{FRT} & \textbf{EMO} & \textbf{DES} & \textbf{INT} & \textbf{KNO} & \textbf{BEL} & \textbf{NLC} \\
        \hline
            \multirow{2}{*}{\textbf{Arabic}} & D & 53.7 & 33.3 & 50.0 & 44.4 & 58.3 & 55.0 & 56.9 & 73.2 & 54.3 & 62.5 & 60.7 & 30.0 & 48.2 & 67.2 \\
        & C & 48.1 & 22.2 & 33.3 & 46.3 & 61.1 & 55.0 & 48.6 & 74.2 & 48.6 & 37.5 & 55.7 & 24.0 & 50.6 & 68.8 \\
        \hline
        \multirow{2}{*}{\textbf{English}} & D & 61.1 & 38.9 & 50.0 & 52.8 & 69.4 & 55.0 & 75.0 & 63.9 & 64.3 & 56.2 & 72.1 & 34.0 & 57.9 & 66.4 \\
        & C & 57.4 & 33.3 & 50.0 & 52.8 & 52.8 & 55.0 & 65.3 & 60.8 & 61.4 & 53.1 & 62.3 & 30.0 & 53.0 & 63.2 \\
        \hline
        \multirow{2}{*}{\textbf{Bengali}} & D & 53.7 & 36.1 & 33.3 & 44.4 & 41.7 & 20.0 & 40.3 & 66.0 & 57.1 & 40.6 & 39.3 & 32.0 & 44.5 & 56.0 \\
        & C & 59.3 & 30.6 & 27.8 & 48.1 & 52.8 & 35.0 & 54.2 & 55.7 & 61.4 & 34.4 & 54.1 & 30.0 & 48.2 & 53.6 \\
        \hline
        \multirow{2}{*}{\textbf{Hindi}} & D & 55.6 & 33.3 & 22.2 & 45.4 & 66.7 & 50.0 & 50.0 & 63.9 & 57.1 & 40.6 & 57.4 & 28.0 & 49.4 & 60.8 \\
        & C & 59.3 & 33.3 & 33.3 & 48.1 & 55.6 & 35.0 & 56.9 & 61.9 & 60.0 & 43.8 & 59.0 & 28.0 & 50.6 & 59.2 \\
        \hline
        \multirow{2}{*}{\textbf{Russian}} & D & 61.1 & 33.3 & 44.4 & 50.0 & 61.1 & 50.0 & 56.9 & 70.1 & 57.1 & 43.8 & 62.3 & 30.0 & 53.7 & 67.2 \\
        & C & 51.9 & 27.8 & 55.6 & 52.8 & 61.1 & 40.0 & 65.3 & 68.0 & 61.4 & 50.0 & 60.7 & 24.0 & 53.7 & 66.4 \\
        \hline
        \multirow{2}{*}{\textbf{French}} & D & 63.0 & 38.9 & 55.6 & 44.4 & 38.9 & 55.0 & 69.4 & 62.9 & 67.1 & 68.8 & 59.0 & 36.0 & 47.6 & 61.6 \\
        & C & 55.6 & 38.9 & 55.6 & 50.0 & 58.3 & 45.0 & 68.1 & 67.0 & 60.0 & 65.6 & 62.3 & 36.0 & 52.4 & 66.4 \\
        \hline
        \multirow{2}{*}{\textbf{Chinese}} & D & 66.7 & 41.7 & 38.9 & 43.5 & 63.9 & 35.0 & 68.1 & 72.2 & 67.1 & 53.1 & 63.9 & 42.0 & 50.0 & 70.4 \\
        & C & 59.3 & 41.7 & 61.1 & 51.9 & 63.9 & 45.0 & 62.5 & 67.0 & 61.4 & 56.2 & 60.7 & 38.0 & 54.9 & 67.2 \\
        \hline

    \hline
    \end{tabular}
    \caption{Performance of Various Tasks and Abilities across Languages and Cultural Contexts (D: Direct/, C: Cultural) for \textbf{Llama-3.1-8b-Instruct}}
    \label{tab:llama3_data_table}
\end{table*}

\begin{table*}[h!]
    \centering
    \small

    \begin{tabular}{|l|l|c|c|c|c|c|c|c|c|c|c|c|c|c|c|}
        \hline
        \textbf{Lang} & & \textbf{UOT} & \textbf{SIT} & \textbf{PST} & \textbf{FBT} & \textbf{AST} & \textbf{HTT} & \textbf{SST} & \textbf{FRT} & \textbf{EMO} & \textbf{DES} & \textbf{INT} & \textbf{KNO} & \textbf{BEL} & \textbf{NLC} \\
        \hline
            \multirow{2}{*}{\textbf{Arabic}} & D & 24.1 & 13.9 & 27.8 & 22.2 & 33.3 & 20 & 15.3 & 35.1 & 22.9 & 25.0 & 21.3 & 22 & 22.6 & 30.4 \\
        & C & 27.8 & 16.7 & 33.3 & 15.7 & 30.6 & 20 & 19.4 & 40.2 & 24.3 & 31.2 & 19.7 & 18 & 19.5 & 35.2 \\
        \hline
        \multirow{2}{*}{\textbf{English}} & D & 31.5 & 16.7 & 27.8 & 24.1 & 13.9 & 5 & 15.3 & 37.1 & 27.1 & 25.0 & 16.4 & 18 & 21.3 & 30.4 \\
        & C & 27.8 & 22.2 & 11.1 & 25.0 & 22.2 & 0 & 13.9 & 41.2 & 27.1 & 12.5 & 16.4 & 18 & 22.0 & 32.8 \\
        \hline
        \multirow{2}{*}{\textbf{Bengali}} & D & 20.4 & 19.4 & 22.2 & 23.1 & 22.2 & 20 & 16.7 & 33.0 & 25.7 & 18.8 & 19.7 & 24 & 19.5 & 29.6 \\
        & C & 24.1 & 19.4 & 27.8 & 14.8 & 30.6 & 15 & 18.1 & 33.0 & 22.9 & 21.9 & 19.7 & 28 & 16.5 & 30.4 \\
        \hline
        \multirow{2}{*}{\textbf{Hindi}} & D & 24.1 & 5.6 & 22.2 & 20.4 & 30.6 & 10 & 23.6 & 24.7 & 21.4 & 25.0 & 27.9 & 12 & 20.1 & 22.4 \\
        & C & 22.2 & 8.3 & 27.8 & 15.7 & 27.8 & 5 & 30.6 & 24.7 & 20.0 & 25.0 & 26.2 & 12 & 18.9 & 23.2 \\
        \hline
        \multirow{2}{*}{\textbf{Russian}} & D & 25.9 & 13.9 & 22.2 & 25.0 & 25.0 & 5 & 16.7 & 41.2 & 25.7 & 21.9 & 18.0 & 14 & 22.0 & 35.2 \\
        & C & 24.1 & 11.1 & 33.3 & 24.1 & 8.3 & 5 & 15.3 & 41.2 & 28.6 & 28.1 & 9.8 & 12 & 18.3 & 35.2 \\
        \hline
        \multirow{2}{*}{\textbf{French}} & D & 24.1 & 19.4 & 38.9 & 23.1 & 25.0 & 10 & 18.1 & 34.0 & 28.6 & 34.4 & 16.4 & 16 & 20.1 & 29.6 \\
        & C & 22.2 & 22.2 & 44.4 & 25.0 & 16.7 & 10 & 13.9 & 34.0 & 22.9 & 34.4 & 14.8 & 18 & 21.3 & 28.0 \\
        \hline
        \multirow{2}{*}{\textbf{Chinese}} & D & 20.4 & 13.9 & 22.2 & 25.0 & 13.9 & 10 & 15.3 & 30.9 & 27.1 & 18.8 & 8.2 & 18 & 22.6 & 25.6 \\
        & C & 27.8 & 16.7 & 22.2 & 26.9 & 27.8 & 15 & 19.4 & 37.1 & 30.0 & 21.9 & 21.3 & 22 & 24.4 & 32.0 \\
        \hline

    \hline
    \end{tabular}
    \caption{Performance of Various Tasks and Abilities across Languages and Cultural Contexts (D: Direct/, C: Cultural) for \textbf{Llama-2-7b-Chat}}
    \label{tab:llama2_data_table}
\end{table*}
    
\clearpage

\clearpage 

\clearpage 

\end{document}